\pdfoutput=1
\documentclass[11pt]{article}

\usepackage{EMNLP2023}

\usepackage{times}
\usepackage{latexsym}
\usepackage{hyperref}
\usepackage{multirow}
\usepackage{amsmath}

\usepackage[T1]{fontenc}
\usepackage[utf8]{inputenc}

\usepackage{microtype}

\usepackage{inconsolata}
\usepackage{graphicx}
\graphicspath{ {./images/} }
\usepackage{booktabs}

\def\github{https://github.com/cmu-llab/gpt_mt_benchmark}
\def\zeno{https://hub.zenoml.com/project/cabreraalex/GPT\%20MT\%20Benchmark}

\newcommand\extrafootertext[1]{%
    \bgroup
    \renewcommand\thefootnote{\fnsymbol{footnote}}%
    \renewcommand\thempfootnote{\fnsymbol{mpfootnote}}%
    \footnotetext[0]{#1}%
    \egroup
}
\usepackage{multirow}
\usepackage{array}
\newcolumntype{H}{>{\setbox0=\hbox\bgroup}c<{\egroup}@{}}
\usepackage{longtable}

\title{ChatGPT MT: Competitive for High- (but not Low-) Resource Languages}

\author{
Nathaniel R. Robinson$^{1,2*}$ 
\quad Perez Ogayo$^{1*}$ 
\quad David R. Mortensen$^{1}$
\quad Graham Neubig$^{1}$
\\
$^1$Language Technologies Institute, Carnegie Mellon University\\
$^2$ Center for Language and Speech Processing, Johns Hopkins University\\
 \texttt{nrobin38@jhu.edu, \{aogayo, dmortens, gneubig\}@cs.cmu.edu}\\
 \small{* Authors contributed equally}\\
}

\begin{document}
\maketitle
% \extrafootertext{* Authors contributed equally}
\begin{abstract}

Large language models (LLMs) implicitly learn to perform a range of language tasks, including machine translation (MT).
Previous studies explore aspects of LLMs' MT capabilities. 
However, there exist a wide variety of languages for which recent LLM MT performance has never before been evaluated. 
Without published experimental evidence on the matter, it is difficult for speakers of the world's diverse languages to know how and whether they can use LLMs for their languages. 
We present the first experimental evidence for an expansive set of 204 languages, along with MT cost analysis, using the FLORES-200 benchmark. 
Trends reveal that GPT models approach or exceed traditional MT model performance for some high-resource languages (HRLs) but consistently lag for low-resource languages (LRLs), under-performing traditional MT for 84.1\% of languages we covered.  
Our analysis reveals that a language's resource level is the most important feature in determining ChatGPT's relative ability to translate it, and suggests that ChatGPT is especially disadvantaged for LRLs and African languages.
\end{abstract}

\section{Introduction}

Despite the  majority of the world's languages being low-resource, current MT systems still perform poorly on them or do not include them at all.
Some commercial systems like Google Translate\footnote{\url{https://translate.google.com}} 
support a number of LRLs, but many systems do not support any, and in either case the majority of LRLs are largely neglected in language technologies.

In recent years, generative LLMs have shown increasingly impressive translation abilities \citep{radford2019language, NEURIPS2020_1457c0d6}. Even more recently, LLM tools like ChatGPT have become popular and accessible to end users. 
This marks an important shift, since a majority of LLM users are now consumers rather than researchers. 
The prospect of LLM translation is exciting, since theoretically, generative LLMs could support more languages than commercial systems like Google's.\footnote{Google Translate currently supports only 133 languages with systems deemed high enough quality for deployment.} 
But only beginning steps have been made to test this hypothesis. 
While some studies outlined in \S\ref{sxn:rw} have evaluated MT with recent LLMs, evaluation is still lacking for many languages. 
This brings up important questions, such as: \emph{Can end users in need of MT for a variety of languages use ChatGPT?} 
\emph{Are ChatGPT and other LLMs reliable translators?} 
\emph{For which languages are they reliable?} Initially we hypothesize that LLMs translate HRLs better than LRLs. 
But due to limited information about the training data and methods for powerful LLMs like ChatGPT (\textbf{GPT-3.5} and variants) and GPT-4, hypotheses like this must be experimentally verified. 

We attempt a significant expansion of experimental verification for such hypotheses by testing ChatGPT's performance on the FLORES-200 benchmark \citep{nllbteam2022language}, containing 204 language varieties.%\nrr{Just a draft here...}
We emphasize that, rather than optimizing LLM MT for a few languages, we focus on helping end users of various language communities know how and when to use LLM MT. 
We expect that our contributions may benefit both direct end users, such as LRL speakers in need of translation, and indirect users, such as researchers of LRL translation considering ChatGPT to enhance specialized MT systems. 
In summary, we contribute:

\begin{enumerate}
    \item MT scores on 203 languages for ChatGPT and comparisons with GPT-4, Google Translate, and NLLB \cite{nllbteam2022language} 
    \item Evidence that LLMs are competitive with traditional MT models for many HRLs but lag for LRLs (with baselines outperforming ChatGPT on 84.1\% of languages evaluated)
    \item Evidence that few-shot prompts offer marginal benefits for LLM translation
    \item A decision tree analysis of language features' correlation with LLM effectiveness in MT, suggesting ChatGPT is especially disadvantaged for LRLs and African languages
    \item A cost comparison across MT systems
\end{enumerate}

Our experiments are motivated by the interests of LLM users speaking a variety of languages. In addition to evaluating a large language set (\S\ref{sxn:res}), we chose to analyse language features (\S\ref{sxn:lang_feats}), to draw generalizations for even more LRL speakers. We compare MT costs because they impact end users (\S\ref{sxn:cost}). We keep ChatGPT central to our analyses because of its current popularity among consumers.

\section{Methodology}
\label{sxn:meth}

We used data for 204 language varieties from FLORES-200 \cite{nllbteam2022language}. We used the 1012 \textit{devtest} sentences for our main experiments and the 997 \textit{dev} sentences for  follow-up  experiments. We queried the OpenAI API\footnote{\url{https://platform.openai.com}} to translate our test set from English into the target languages. We explored ENG$\rightarrow$X translation only because the FLORES-200 English data was taken from Wikipedia. Thus OpenAI's GPT models were likely trained on those exact English sentences, making fair X$\rightarrow$ENG evaluation infeasible. 

\subsection{Experimental setup}
\label{sxn:exp_setup}

We evaluated ChatGPT's (\texttt{gpt-3.5-turbo}) MT for our full language set. 
We  compared with NLLB-MOE \cite{nllbteam2022language} as our baseline, as it is the current state-of-the-art open-source MT model that covers such a wide variety of languages. 
NLLB is a discriminative transformer trained on supervised bi-text data (the traditional MT paradigm). We obtained scores for NLLB outputs of ENG$\rightarrow$X translation into 201 of the language varieties in our set (as reported by \citet{nllbteam2022language}).

We used both zero- and five-shot prompts for ChatGPT MT. 
(See \S\ref{sxn:fewshot}.) 
Previous studies \cite{hendy2023good, gao2023design, moslem2023adaptive, NEURIPS2020_1457c0d6, zhu2023multilingual} suggest that few-shot prompts produce slightly (albeit not consistently) better translations. 
But zero-shot prompts are more convenient and affordable for users.  

We also compare with results for subsets of our selected languages from two other MT engines. 
Google Translate API was an important baseline for our analysis because it is popular among end users. 
We also included it to represent commercial MT systems in our study. 
Because Google's API does not support all 204 of the FLORES-200 languages, we obtained results only for the 115 non-English languages it supports. 

Lastly, we obtained MT results from GPT-4, since it is a popular LLM and has been shown to outperform ChatGPT on MT \cite{jiao2023chatgpt, wang2023document}. 
Because the cost of GPT-4 use exceeds that of ChatGPT by 1900\%, our resources did not permit its evaluation on all 203 non-English languages. 
Instead we selected a 20-language subset by picking approximately every 10th language, with languages sorted by chrF++ differentials between ChatGPT and NLLB ($chrf_{GPT} - chrf_{NLLB}$). 
We chose this criterion in order to have 20 languages with a range of relative ChatGPT performance and a variety of resource levels. 
We used only five-shot prompts for GPT-4. 

\subsection{Implementation details}
\label{sxn:implement}

We conducted all LLM experiments with \texttt{gpt-3.5-turbo} (ChatGPT) and \texttt{gpt-4-0613} (GPT-4). We used \texttt{top\_p} $1$, \texttt{temperature} $0.3$, \texttt{context\_length} $-1$, and \texttt{max\_tokens}\footnote{Although some languages had higher token counts than others (see \S\ref{sxn:lang_feats}), we found that adjusting \texttt{max\_tokens} had a minimal effect on MT performance. We thus decided to maintain the same value of \texttt{max\_tokens} across all languages for experimental consistency.} $500$.

To evaluate the outputs, we used:\footnote{We excluded learned MT metrics like COMET \cite{rei2020comet} and BLEURT \cite{sellam2020bleurt}, since they do not support many LRLs.}

\noindent\textbf{spBLEU}: BLEU \cite{bleu-paper} is standard in MT evaluation. We find spBLEU scores \cite{goyal-etal-2022-flores} via sacreBLEU \cite{post-2018-call} with the SPM-200 tokenizer \cite{nllbteam2022language}.

\noindent\textbf{chrF2++}: We  use sacreBLEU's implemantation of  chrF++ \cite{popovic-2017-chrf}. We adopt it as our main metric, as it overcomes some of BLEU's weaknesses, and refer to it as \textit{chrF} for brevity.

\subsection{Zero- and few-shot prompts}
\label{sxn:fewshot}

Previous works \cite{gao2023design, jiao2023chatgpt} investigated LLM prompting to optimize MT performance. 
We adopt \citet{gao2023design}'s recommended prompts for both zero- and few-shot MT (Table~\ref{tab:prompts}). 
We are interested in multiple $n$-shot prompt settings because, as mentioned in \S\ref{sxn:exp_setup}, they present different benefits to LLM users. 
We explored zero-shot (no in-context example), one-shot (1 example), and five-shot (5 examples). 
We employed both zero- and five-shot prompts in our main experiments over 203 languages, and we analyzed all three $n$-shot settings for a subset of languages on FLORES-200 \textit{dev} sets. 

\begin{table}[t]
\small
\begin{tabular}{p{0.04\linewidth}p{0.90\linewidth}}
Shot & Prompt  \\
\hline
zero     & This is an English to {[}TGT{]} translation, please provide the {[}TGT{]} translation for this sentence. Do not provide any explanations or text apart from the translation. \newline{[}SRC{]}: [src-sentence]
\newline {[}TGT{]}:                                                                                                       \\

five     & This is an English to {[}TGT{]} translation, please provide the {[}TGT{]} translation for these sentences: 

{[}SRC{]}: [src-sentence]
{[}TGT{]}: [tgt-sentence]

{[}SRC{]}: [src-sentence]
{[}TGT{]}: [tgt-sentence]

{[}SRC{]}: [src-sentence]
{[}TGT{]}: [tgt-sentence]

{[}SRC{]}: [src-sentence]
{[}TGT{]}: [tgt-sentence]

{[}SRC{]}: [src-sentence]
{[}TGT{]}: [tgt-sentence]

Please provide the translation for the following sentence. Do not provide any explanations or text apart from the translation. \newline{[}SRC{]}: [src-sentence]
\newline {[}TGT{]}:
  \\
\end{tabular}
\caption{%\centering
Prompts used for zero- and five-shot settings}
\label{tab:prompts}
\end{table}

The languages in FLORES-200 represent 22 language families. 
To experiment with multiple $n$-shot settings, we selected one language from each of the 12 families containing at least two members in the set. 
We chose four HRLs ($\geq$1M Wikipedia pages\footnote{Throughout the paper we use the "Total pages" count from \url{https://en.wikipedia.org/wiki/List_of_Wikipedias}, accessed 7 August 2023, as a proxy for the resource level of a language.}), four  LRLs (25K-1M pages), and four extremely LRLs ($\leq$25K pages).  
These languages also employ a variety of scripts. 
See Table~\ref{tab:fewshotlangs}. 

\begin{table}[ht]
\footnotesize
\centering
\begin{tabular}{lcccr}
\textbf{Language} & \textbf{Code} & \textbf{Family} & \textbf{Script} & \textbf{Wiki. \#}\\
\hline
French & \texttt{fra} & Indo-European & \texttt{Latn} & 12.7M \\
Chinese & \texttt{zho} & Sino-Tibetan & \texttt{Hans} & 7.48M \\
Turkish & \texttt{tur} & Turkic & \texttt{Latn} & 2.48M \\
Finnish & \texttt{fin} & Uralic & \texttt{Latn} & 1.46M \\
Tamil & \texttt{tam} & Dravidian & \texttt{Taml} & 496K \\
Tagalog & \texttt{tgl} & Austronesian & \texttt{Latn} & 239K \\
Kiswahili & \texttt{swh} & Niger-Congo & \texttt{Latn} & 167K \\
Amharic & \texttt{amh} & Afroasiatic & \texttt{Ethi} & 46.2K \\
Santali & \texttt{sat} & Austroasiatic & \texttt{Olck} & 20.0K \\
Lao & \texttt{lao} & Kra-Dai & \texttt{Laoo} & 14.0K \\
Papiamento & \texttt{pap} & Creole & \texttt{Latn} & 6.84K \\
Luo & \texttt{luo} & Nilo-Saharan & \texttt{Latn} & 0 
\end{tabular}
\caption{%\centering
Diverse subset of languages experiments with few-shot settings. \textbf{Wiki. \#} is the number of Wikipedia pages in the language.}
\label{tab:fewshotlangs}
\end{table}

\section{Results and Analysis}
\label{sxn:res}

\subsection{Traditional MT generally beats LLMs}

Table~\ref{tab:avgs} shows the number of languages we evaluated for each MT system, as noted in \S\ref{sxn:exp_setup}, with average chrF and BLEU scores across those languages. 
The best performing model on average was (1) Google, then (2) NLLB, (3) GPT-4, and (4) ChatGPT. 
Unabridged results are in Table~\ref{tab:main_results} in Appendix~\ref{sec:appendix}. Supplementary materials can also be browsed on our \href{\github}{repository}.\footnote{\url{\github}} 
(Also see the interactive score visualizer on our Zeno \href{\zeno}{browser}.\footnote{\url{\zeno}})  

Table~\ref{tab:gpt-4_res} shows a meaningful subset of scores: chrF for the 20 languages evaluated on both LLM systems. 
Of the 11 languages evaluated on all four systems, Google performed best for 10 of them. 
Notably, GPT-4 surpassed NLLB in five languages and Google in one (Moroccan Arabic, \texttt{acm\_Arab}).

\begin{table}[]
    \centering
    \small
    \begin{tabular}{lccc}
         & & \textbf{avg.} & \textbf{avg.} \\
         & \textbf{\#langs.} & \textbf{chrF} & \textbf{BLEU} \\
         \hline
         \textbf{ChatGPT} (0-shot) & 203 & 32.3 & 16.7 \\
         \textbf{ChatGPT} (5-shot) & 203 & 33.1 & 17.3 \\
         \textbf{GPT-4} & 20 & 44.6 & 24.6 \\
         \textbf{NLLB} & 201 & 45.3 & 27.1 \\
         \textbf{Google} & 115 & \textbf{52.2} & \textbf{34.6} 
    \end{tabular}
    \caption{Languages evaluated, average chrF, and average BLEU for each MT system. Best scores are \textbf{bold}.}
    \label{tab:avgs}
\end{table}

\begin{table}[ht]
\small
\centering
\begin{tabular}{lHHHcccc}
\textbf{Lang.}     & Resource & spBLEU100 & \textbf{BLEU} & \textbf{GPT-4} & \multicolumn{1}{c}{\textbf{ChatGPT}} & \multicolumn{1}{c}{\textbf{Google}}  & \textbf{NLLB}\\
\hline
ssw\_Latn & low      & 6.5                          & 5.8                          & 24.1                    & 6.7                            & \multicolumn{1}{c}{-}                   & 43.3                         \\
sna\_Latn & low      & 10.0                         & 8.4                          & 29.2                    & 16.3                           & \textbf{44.4} & 43.4                          \\
ckb\_Arab & low      & 12.9                         & 11.2                         & 33.1                    & 24.8                           & \textbf{47.7}                         & 47.2                          \\
mag\_Deva & low      & 27.6                         & 24.8                         & 44.6                    & 39.9                           & \multicolumn{1}{c}{-}                   & 58.5                          \\
ibo\_Latn & low      & 11.5                         & 9.8                          & 27.7                    & 16.3                           & \textbf{43.5}                         & 41.4                          \\
hau\_Latn & low      & 16.6                         & 15.7                         & 40.3                     & 22.4                           & \textbf{53.2}                         & 53.5                          \\
pbt\_Arab & low      & 9.0                          & 9.2                          & 26.7                     & 21.1                           & \multicolumn{1}{c}{-}                   & 39.4                          \\
tam\_Taml & low      & 20.8                         & 20.9                         & 42.7                    & 34.5                           & \textbf{55.8}                         & 53.7                          \\
kat\_Geor & low      & 21.2                         & 23.2                         & 41.4                    & 33.5                           & \textbf{51.4}                         & 48.1                          \\
gle\_Latn & low      & 36.9                         & 32.8                         & 53.0                    & 47.5                           & \textbf{60.1}                         & 58.0                            \\
kmr\_Latn & low      & 16.0                         & 14.3                         & 34.3                    & 27.4                           & \textbf{40.0}                         & 39.3                          \\
war\_Latn & low      & 29.9                         & 28.5                         & 54.0                    & 49.5                           & \multicolumn{1}{c}{-}                   & 57.4                          \\
ajp\_Arab & low      & 27.3                         & 32.2                         & 48.4                    & 47.5                           & \multicolumn{1}{c}{-}                   & 51.3                          \\
lim\_Latn & low      & 20.8                         & 21.0                         & 45.1                    & 42.7                           & \multicolumn{1}{c}{-}                   & 47.9                          \\
ukr\_Cyrl & High     & 37.9                         & 39.2                         & 56.3                    & 55.4                            & \textbf{58.6}                          & 56.3                          \\
fra\_Latn & High     & 56.6                         & 57.3                          & 71.7                    & 71.3                           & \textbf{72.7}                         & 69.7                          \\
lvs\_Latn & High     & 37.2                         & 36.7                         & 57.3                    & 55.2                           & \multicolumn{1}{c}{-}                   & 54.8                          \\
ron\_Latn & High     & 47.2                          & 49.0                         & \textbf{65.3}           & 64.2                           & 65.0                                  & 61.3                          \\
tpi\_Latn & low      & 26.7                         & 22.7                         & \textbf{49.5}                    & 39.2                           & \multicolumn{1}{c}{-}                   & 41.6                          \\
acm\_Arab & low      & 25.3                         & 29.5                         & \textbf{46.5}                    & 46.1                           & \multicolumn{1}{c}{-}                   & 31.9      \\
\hline
\end{tabular}
\caption{%\centering
chrF ($\uparrow$) scores across models for all languages we used to evaluate GPT-4. Best scores are \textbf{bold}. ChatGPT scores here are 5-shot, to compare with GPT-4.
}
\label{tab:gpt-4_res}
\end{table}

On the 20 languages for which we tested it, GPT-4 improved over ChatGPT by 6.5 chrF on average. 
The standard deviation of performance difference with NLLB ($chrF_{GPT} - chrF_{NLLB}$) was 8.6 for GPT-4, compared with ChatGPT's 12.7 for the same languages, suggesting a more consistent advantage across language directions.  
GPT-4 offered larger improvements for LRLs, whereas HRL performance plateaued between the LLMs. 
Previous studies have found GPT-4 improving multilingual capabilities over ChatGPT on a range of tasks \cite{xu2023superclue, zhang2023m3exam,openai2023gpt4}. 
This may account for its superior MT performance. 

Google Translate outperformed all other systems in chrF on 100 of the 115 languages for which we evaluated it, with an average improvement of 2.0 chrF points over the next best system for each language. 
(See Appendix~\ref{sec:appendix} for unabridged results.) 
Google's was the best performing MT system overall, though NLLB has broader language coverage. 

NLLB outperformed ChatGPT in chrF on 169 (84.1\%) of the 201 languages for which we obtained scores for both, with NLLB scoring an average of 11.9 chrF points higher than the better $n$-shot ChatGPT setting for each language. 
This trend is corroborated by \citet{zhu2023multilingual}. 
Table~\ref{tab:outliers-zero-shot} has both BLEU and chrF scores from both systems for the five languages with the most negative chrF deltas ($chrF_{GPT} - chrF_{NLLB}$) on top, followed by the five languages with the highest positive deltas on bottom. 
For many of the subsequent sections of this paper we focus on comparing ChatGPT and NLLB, since we evaluted them on the most languages. 

\begin{table}[ht]
\small
\centering
\begin{tabular}{l|rr|rr}

& \multicolumn{2}{c}{ChatGPT} & \multicolumn{2}{|c}{NLLB}\\
\textbf{Lang.} & \multicolumn{1}{l}{\textbf{BLEU}} & \multicolumn{1}{l|}{\textbf{chrF}} & \multicolumn{1}{c}{\textbf{BLEU}} & \multicolumn{1}{l}{\textbf{chrF}} \\
\hline
srp\_Cyrl     & 1.36    & 3.26       &  \textbf{\textcolor{blue}{43.4} }                                  &  \textbf{59.7}                                \\
kon\_Latn     & 0.94    & 8.50     & \textbf{\textcolor{blue}{18.9}}                                   & \textbf{45.3}                                 \\
tso\_Latn     & 2.92     & 15.0  & \textbf{\textcolor{blue}{26.7}}                                   & \textbf{50.0}                                   \\
kac\_Latn     & 0.04     & 2.95    & \textbf{\textcolor{blue}{14.3}}                                   & \textbf{37.5}                                 \\
nso\_Latn     & 3.69   & 16.7     & \textbf{\textcolor{blue}{26.5}}                                   & \textbf{50.8}                                 \\
% lin\_Latn     & 2.59    & 14.8    & \textbf{\textcolor{blue}{21.9}}                                   & \textbf{48.0}                                   \\
% ewe\_Latn     & 0.70    & 5.97    & \textbf{\textcolor{blue}{17.2}}       & \textbf{39.0}       \\
% ssw\_Latn     & 1.90    & 10.6      & \textbf{\textcolor{blue}{19.9}}      & \textbf{43.3}                                 \\
% tsn\_Latn     & 4.00                                      & 16.7                                & \textbf{\textcolor{blue}{25.6}}                                   & \textbf{48.5}                                 \\
% hau\_Latn     & 6.29   & 21.9                                & \textbf{\textcolor{blue}{31.4}}                                  & \textbf{53.5}                                 \\
\midrule
% spa\_Latn     & \textbf{\textcolor{blue}{33.8}}    & \textbf{56.6}   & 33.1                                   & 53.8                                 \\
% aeb\_Arab     & \textbf{\textcolor{blue}{24.2}}     & \textbf{41.0}                                & 19.9   & 38.2                                 \\
% por\_Latn     & \textbf{\textcolor{blue}{56.4}}    & \textbf{71.4}    & 52.9   & 67.9                                 \\
% yue\_Hant     & \textbf{\textcolor{blue}{26.4}}   & \textbf{22.3}                                 & 16.6                                   & 17.9                                 \\
% acq\_Arab     & \textbf{\textcolor{blue}{29.3}}  & \textbf{47.0}                                & 26.9                                   & 42.2                                 \\
jpn\_Jpan     & \textbf{\textcolor{blue}{28.4}}   & \textbf{32.9}  & 20.1                                   & 27.9                                 \\
nno\_Latn     & \textbf{\textcolor{blue}{37.1}}  & \textbf{58.7}                                & 33.4 & 53.6                                 \\
zho\_Hans     & \textbf{\textcolor{blue}{36.3}}                                  & \textbf{31.0}                                & 26.6  & 22.8                                 \\
zho\_Hant     & \textbf{\textcolor{blue}{26.0}}                                  & \textbf{24.4}                                & 12.4   & 14.0                                   \\
acm\_Arab     & \textbf{\textcolor{blue}{28.2}}                                  & \textbf{44.7}                                & 11.8  & 31.9                              
\end{tabular}
\caption{%\centering
Lowest (top) and highest (bottom) chrF differences between zero-shot ChatGPT and NLLB. Best scores for each metric in \textbf{bold} (with BLEU \textbf{\textcolor{blue}{blue}}).
}
\label{tab:outliers-zero-shot}
\end{table}

\subsection{ChatGPT under-performs for LRL}

Using \citeauthor{nllbteam2022language}'s (\citeyear{nllbteam2022language}) resource categorization, we find that ChatGPT performs worse on LRLs than HRLs, corroborating findings of previous works \cite{jiao2023chatgpt, zhu2023multilingual}. 
There is a strong positive correlation between ChatGPT and NLLB chrF scores, but the correlation is higher for HRLs ($\rho$=0.85) than LRLs ($\rho$=0.78), indicating that ChatGPT struggles to keep up with NLLB for LRLs.

\begin{figure*}
    \centering
    \includegraphics[width=0.86\linewidth]{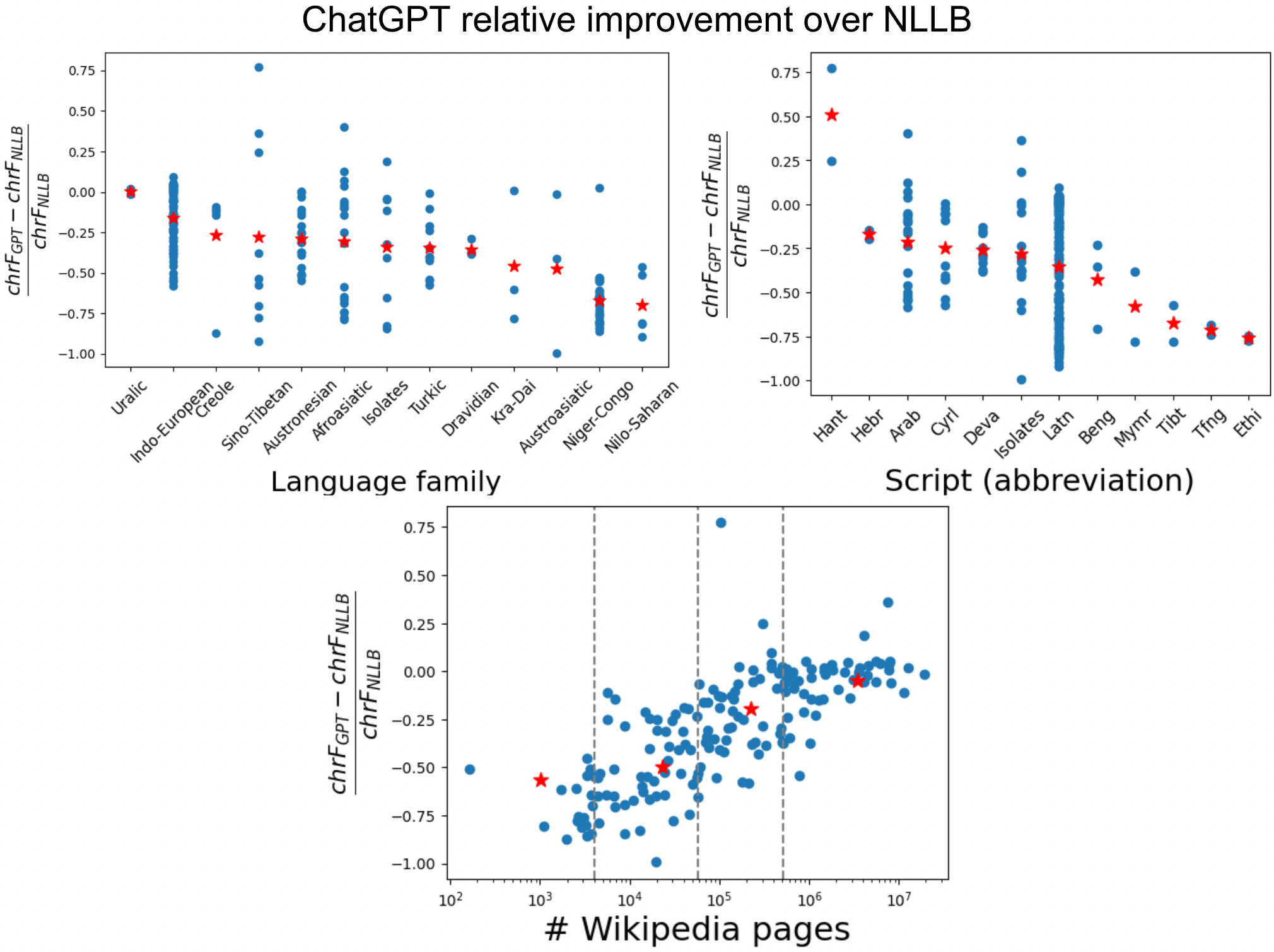}
    \caption{%\centering
    ChatGPT \textit{relative improvement} over NLLB chrF, with languages organized by family, script, and number of Wikipedia pages. Red stars represent averages per group. In the bottom plot, languages are grouped into quartiles of equal size (with dotted lines at the Q1, median, and Q3). More expansive visualizations with language labels for each value can be found in Appendix~\ref{sxn:app_hexbins}.
    }
    \label{fig:scatterplots}
\end{figure*}

Figure~\ref{fig:scatterplots} shows scatter plots where dots represent languages, with ChatGPT's (positive or negative) \textit{relative improvement} over NLLB chrF ($\frac{chrf_{GPT} - chrf_{NLLB}}{chrf_{NLLB}}$) on the y-axis. 
When languages are grouped by family or script, some trends are apparent (in part because we ordered groups by descending average scores). 
For example, ChatGPT fairs better with Uralic and Indo-European languages and clearly worse with Niger-Congo and Nilo-Saharan languages. 
However, the clearest natural correlation appears when languages are grouped by resource level, approximated by number of Wikipedia pages (Figure~\ref{fig:scatterplots}, bottom). 
Note the \textit{relative improvement} (y-axis) is typically negative since ChatGPT rarely outperformed NLLB.

In the five-shot setting, ChatGPT outperformed NLLB  on 47\% of the HRLs designated by \citet{nllbteam2022language}, but only on 6\% of the LRLs. 
These findings contrast with what is commonly observed in multilingual MT models \cite{liu-etal-2020-multilingual-denoising, m2m-paper, siddhat-synergy, bapna2022building, nllbteam2022language}, where LRLs benefit the most. 
This  highlights the need to investigate how decoder-only models may catch up with encoder-decoder models in low-resource applications. 
It underscores the importance of smaller specialized models when large multitask models cannot overcome low-resource challenges.

\subsection{Few-shot prompts offer marginal improvement}

Our main experiments suggested that $n$-shot setting had only a modest effect on MT performance. 
We conducted a more concentrated study of $n$-shot prompts using $dev$ sets for the 12 languages in Table~\ref{tab:fewshotlangs}.
Results in Table~\ref{tab:fewshotres} show five-shot prompts performing best.  
For some LRLs, this was simply a result of ChatGPT's failure to model the language. 
In Santali's case, for example, zero-shot ChatGPT was unable to produce the Ol Chiki script at all. 
In the five-shot setting, it was able to imitate the script characters from the context, but without any coherence or accuracy.  
Excepting Santali as an outlier, five-shot settings offered generally marginal improvements over zero-shot (the most cost-effective of the settings), with an average improvement of only 1.41 chrF across all 12 languages (0.31 if we exclude Santali). 
Zero-shot prompts actually produced the best chrF score for six of the 12 languages. 
The one-shot setting performed worst. 
We noted this trend of few-shot contexts offering only meager and inconsistent improvements throughout our experiments, with five-shot MT improving on zero-shot by only 0.88 average chrF across all 203 language directions. (See Appendix~\ref{sec:appendix}.)

\begin{table}[ht]
\small
\centering
\begin{tabular}{l|rr|rr|rr}

& \multicolumn{2}{c}{0-shot} & \multicolumn{2}{|c}{1-shot} & \multicolumn{2}{|c}{5-shot} \\
% \hline
& \multicolumn{1}{l}{\textbf{\textsc{bleu}}} & \multicolumn{1}{l|}{\textbf{chrF}} & \multicolumn{1}{c}{\textbf{\textsc{bleu}}} & \multicolumn{1}{l|}{\textbf{chrF}} & \multicolumn{1}{c}{\textbf{\textsc{bleu}}} & \multicolumn{1}{l}{\textbf{chrF}} \\
\hline
\texttt{fra} & 55.4 & \textbf{71.3} & 50.4 & 70.3 & 55.4 & 71.2 \\
\texttt{zho} & 30.0 & 29.9 & 28.2 & 30.8 & \textbf{\textcolor{blue}{30.7}} & \textbf{31.1} \\
\texttt{fin} & 34.6 & 56.6 & 31.7 & 56.3 & 34.6 & \textbf{56.7} \\
\texttt{tur} & 38.2 & \textbf{58.6} & 34.8 & 57.6 & \textbf{\textcolor{blue}{38.3}} & 58.6 \\
\texttt{tgl} & 35.9 & \textbf{60.2} & 35.2 & 59.6 & \textbf{\textcolor{blue}{36.1}} & 60.1 \\
\texttt{tam} & \textbf{\textcolor{blue}{13.8}} & \textbf{35.3} & 11.7 & 34.3 & 11.9 & 34.6 \\
\texttt{swh} & 39.7 & \textbf{60.6} & 36.0 & 59.5 & \textbf{\textcolor{blue}{40.0}} & 60.5 \\
\texttt{amh} & 3.4 & 10.1 & 3.2 & 9.6 & \textbf{\textcolor{blue}{3.9}} & \textbf{10.6}\\
\texttt{pap} & 26.6 & 51.5 & 29.3 & 54.1 & \textbf{\textcolor{blue}{34.8}} & \textbf{56.1} \\
\texttt{lao} & 4.8 & 21.6 & 4.4 & 20.8 & \textbf{\textcolor{blue}{5.3}} & \textbf{22.1} \\
\texttt{luo} & \textbf{\textcolor{blue}{0.8}} & \textbf{7.6} & 0.2 & 4.6 & 0.2 & 5.2 \\
\texttt{sat} & 0.0 & 0.3 & 2.2 & 11.3 & \textbf{\textcolor{blue}{3.0}} & \textbf{13.8}                              
\end{tabular}
\caption{%\centering
Three $n$-shot settings for 12 diverse languages
}
\label{tab:fewshotres}
\end{table}

\subsection{Importance of language features}
\label{sxn:lang_feats}

We were interested in which language features determined LLMs' effectiveness compared to traditional MT. 
Analyzing this may reveal trends helpful to end users deciding which MT system to use, especially if their language is not represented here but shares some of the features we consider. 
In this section we focus on comparing ChatGPT and NLLB, since we evaluated the most languages with them. 
We focus on zero-shot ChatGPT, as it is the most common and convenient setting for end users. 

We encoded each of the 203 languages in our set as a \textit{feature vector}. In these language \textit{feature vectors} we included \textbf{four numerical features}: number of Wikipedia pages in the language (\texttt{wiki\_ct}), size of the language's bi-text corpus in the Oscar MT database\footnote{\url{https://oscar-project.org}} (\texttt{oscar\_ct}) \cite{oscar2022}, percentage of ASCII characters\footnote{Percentage of characters with an encoding between 0 and 128, inclusive, using the Python built-in \texttt{ord} function} in the FLORES-200 \textit{dev} set for the language (\texttt{ascii\_percentage}), and average number of tokens per \textit{dev} set sentence in FLORES-200 with ChatGPT's tokenizer (\texttt{token\_ct}). 
We also included \textbf{two categorical features}: language family (\texttt{family}) and script the language was written in (\texttt{script}); and \textbf{one binary feature}: the FLORES resource designation of the language--with 1 for high-resource and 0 for low-resource (\texttt{hi/lo}). 
Before analysis, we one-hot encoded the two \textbf{categorical features} into 48 binary features like \texttt{family\_Niger-Congo} and \texttt{script\_Latn}.

We selected \texttt{token\_ct} as a feature because we observed languages in low-resource scripts having many tokens. 
For example, ChatGPT's tokenizer encodes multiple tokens for every character in Ol Chiki script. 
This tendency for GPT models with low-resource scripts has been noted in previous studies \cite{ahia2023languages}. 

\begin{figure*}
    \centering
    \includegraphics[width=\linewidth]{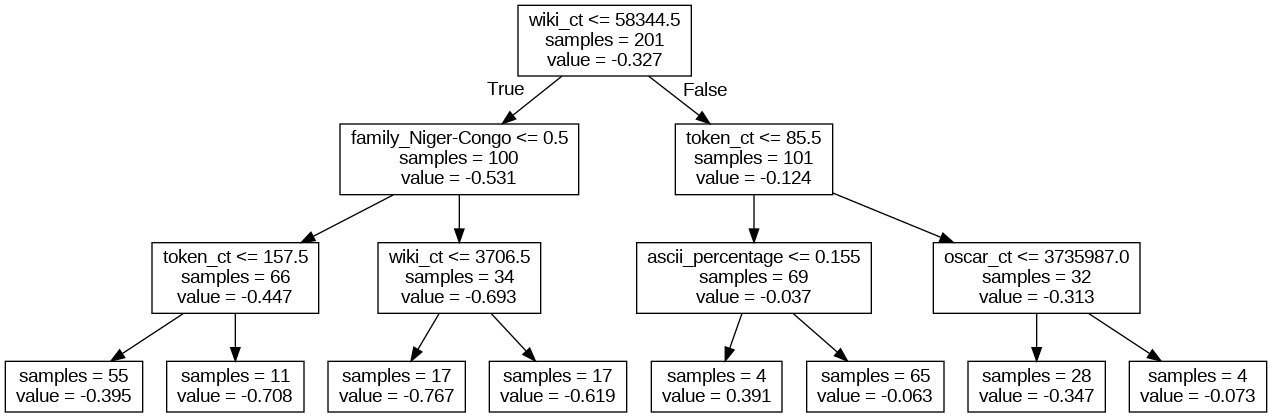}
    \caption{Decision tree predicting ChatGPT \textit{relative improvement} over NLLB chrF, from language features.}
    \label{fig:tree}
\end{figure*}

We fit a decision tree with these \textit{feature vectors} to regress on ChatGPT's \textit{relative improvement} over NLLB in chrF  ($\frac{chrf_{GPT} - chrf_{NLLB}}{chrf_{NLLB}}$), for each of the 201 languages with NLLB scores. 
%(I.e. ChatGPT's percentage improvement over NLLB in chrF2++.) 
When we used \texttt{max\_depth} 3, the tree in Figure \ref{fig:tree} was learned. 
Languages are delimited first by \texttt{wiki\_ct}; then LRLs are separated into Niger-Congo languages and others, while HRLs are delimited by \texttt{token\_ct}. 
The only group where ChatGPT beat NLLB is of languages with more than 58,344 Wikipedia pages, fewer than 86 tokens per average sentence, and less than 15.5\% ASCII characters. 
This group contains some East Asian HRLs. 
The group where ChatGPT was least advantaged contains Niger-Congo languages with fewer than 3,707 Wikipedia pages. 

We also fit a random forest regressor with the same features and labels to find feature importance values. 
Only ten features had importance $\geq0.01$, shown in Table \ref{tab:feat_imps}. 
The most important feature by far was \texttt{wiki\_ct}. 
(This feature correlates strongly with ChatGPT's \textit{relative improvement}, $\rho=0.68$.) 
\texttt{family\_Niger-Congo} was much more important than any other family feature. 
No script feature had an importance exceeding $0.01$. 
In general, features for resource level and tokenization were more important than family or script. 

ChatGPT has a blind spot not only for Niger-Congo languages, but for African languages in general. 
Figure~\ref{fig:scatterplots} shows ChatGPT is least advantaged for the two exclusively African families, Niger-Congo and Nilo-Saharan; and the two exclusively African scripts, Tifinagh (\texttt{Tfng}) and Ge'ez (\texttt{Ethi}). 

\begin{table}[]
\small
    \centering
    \begin{tabular}{lr}
        \textbf{feature} &  \textbf{importance} \\
        \hline
        \texttt{wiki\_ct} & 0.514 \\
        \texttt{token\_ct} & 0.157 \\
        \texttt{ascii\_percentage} & 0.104 \\
        \texttt{family\_Niger-Congo} & 0.054 \\
        \texttt{oscar\_ct} & 0.040 \\
        \texttt{family\_Afroasiatic} & 0.025 \\
        \texttt{family\_Indo-European} & 0.025 \\
        \texttt{family\_Sino-Tibetan} & 0.022 \\
        \texttt{family\_Creole} & 0.012 \\
        \texttt{family\_Nilo-Saharan} & 0.011
    \end{tabular}
    \caption{%\centering
    Ten most important language features to predict ChatGPT's effectiveness relative to NLLB 
    }
    \label{tab:feat_imps}
\end{table}

\subsection{Impact of script}
\label{sxn:script_comp}

Prior research suggests that ChatGPT output quality is sensitive to language script \cite{bang2023multitask}. 
Our own analysis in \S\ref{sxn:lang_feats} actually suggests that script is the least important language feature in predicting ChatGPT's MT effectiveness. 
However, differences in performance are clear when comparing scripts used for the same language. 
Table \ref{tab:mult_scripts} shows one script typically outperforming the other, by an average of 14.3 chrF points for zero-shot. 
Five-shot contexts narrowed the gap slightly to 12.0.  
Although transliteration is a deterministic process for many languages, these performance gaps suggest that ChatGPT has not implicitly learned it as part of a translation task. 
We hypothesize that ChatGPT's observed sensitivity to script in earlier studies may be particular to the languages and tasks evaluated.

\begin{table}[ht]
\small
\centering
\begin{tabular}{l|rr|rr}
          & \multicolumn{2}{c|}{\textbf{BLEU}}                             & \multicolumn{2}{c}{\textbf{chrF}}                               \\
\textbf{Lang.}      & \multicolumn{1}{l}{0-shot} & \multicolumn{1}{l|}{5-shot} & \multicolumn{1}{l}{0-shot} & \multicolumn{1}{l}{5-shot} \\
\hline
ace\_Arab & 1.27                        & 2.26                        & 8.41                        & 9.75                        \\
\textbf{ace\_Latn} & \textbf{\textcolor{blue}{4.98}}                        & 4.35                        & \textbf{19.82}                       & 17.96                       \\
\hline
\textbf{arb\_Arab} & 37.60                        & \textbf{\textcolor{blue}{37.85}}                       & 53.79                       & \textbf{53.81}                       \\
arb\_Latn & 5.33                        & 8.38                        & 22.79                       & 26.92                       \\
\hline
bjn\_Arab & 1.96                        & 3.05                        & 10.43                       & 13.24                       \\
\textbf{bjn\_Latn} & 10.96                       & \textbf{\textcolor{blue}{12.29}}                       & 35.92                       & \textbf{37.98}                       \\
\hline
\textbf{kas\_Arab} & \textbf{\textcolor{blue}{3.99}}                       & 3.30                         & \textbf{15.51}                       & 14.33                       \\
kas\_Deva & 2.31                        & 2.68                        & 12.91                       & 13.91                       \\
\hline
knc\_Arab & 0.51                        & 1.06                        & 5.26                        & 4.67                        \\
\textbf{knc\_Latn} & \textbf{\textcolor{blue}{2.61}}                        & 0.91                        & \textbf{13.38}                       & 8.11                        \\
\hline
min\_Arab & 1.56                        & 3.49                        & 10.06                       & 14.88                       \\
\textbf{min\_Latn} & 11.51                       & \textbf{\textcolor{blue}{13.07}}                       & 36.99                       & \textbf{38.43}                       \\
\hline
taq\_Latn & 0.82                        & 0.28                        & 8.18                        & 6.24                        \\
\textbf{taq\_Tfng} & 0.62                        & \textbf{\textcolor{blue}{1.37}}                        & 5.23                        & \textbf{8.31}                        \\
\hline
\textbf{zho\_Hans} & 36.33                       & \textbf{\textcolor{blue}{36.51}}                       & 31.03                       & \textbf{31.89}                       \\
zho\_Hant & 29.30                        & 30.38                       & 24.82                       & 26.02      \\
% \hline
\end{tabular}
\caption{%\centering
ChatGPT performance on languages with multiple scripts. Each better scoring script is \textbf{bold}.}
\label{tab:mult_scripts}
\end{table}

\subsection{LLMs often get the language wrong}

LLMs' performing worse than NLLB may be due in large part to their translating into the wrong language. 
Using FLORES-200's \textit{dev} data, we trained a logistic regression language identifier for 100 epochs. 
Language identification accuracies for four of the models we evaluated are in Table~\ref{tab:langID}. 
Zero-shot ChatGPT only translated on target 72\% of the time. 
This expectedly improved with five-shot prompts, and GPT-4 performed even better, still just shy of NLLB. 
LLMs' tendency to translate off target is corroborated by \citet{zhu2023multilingual}.

\begin{table}[ht]
    \centering
    \small
    \begin{tabular}{lr}
        \textbf{model} &  \textbf{lang. ID acc.} \\
        \hline
        ChatGPT (0-shot) & 72\% \\
        ChatGPT (5-shot) & 83\% \\
        GPT-4 (5-shot) & 90\% \\
        NLLB & 91\% \\
    \end{tabular}
    \caption{%\centering
    Proportion of the time each model translated into the correct target language}
    \label{tab:langID}
\end{table}

\subsection{Cost comparison}
\label{sxn:cost}

Our results suggest that GPT-4 is a better translator than ChatGPT. 
However in considering the needs of MT end users, it would be remiss not to consider the respective costs of the systems evaluated. 
GPT-4's high cost (roughly 2000\% that of ChatGPT's) prohibited us from evaluating it on all FLORES-200 languages. 
In general, using few-shot prompts for LLMs is more costly than zero-shot prompts, since users are charged for both input and output tokens. 
And for this same reason, some languages are more costly than others in LLM MT. 
Previous work has found that Google Translate has associated costs comparable to those of five-shot ChatGPT \cite{Neubig_Zeno_GPT_Machine_2023}. 
NLLB is the least expensive system we evaluated. 

We estimated cost values for each MT system and language: the expense, in USD, of translating the full FLORES-200 \textit{devtest} English set into the language. 
We estimated costs of GPT models using the prompts employed in our experiments, the tiktoken tokenizer\footnote{\url{https://github.com/openai/tiktoken}} used by both models, and inference prices from OpenAI's website.\footnote{\url{https://openai.com/pricing}} 
Conveniently, Google Translate costs nothing for the first 500K input characters. 
But since frequent MT users may have already expended this allowance, we calculated costs from their rates beyond the first 500K.\footnote{\url{https://cloud.google.com/translate/pricing}} 
As the full NLLB-MOE model (54.5B parameters) is difficult to run on standard computing devices, \citet{nllbteam2022language} also provided a version with only 3.3B parameters that achieves similar performance. 
Since users commonly opt for the smaller model, and since the performance difference does not impact our estimates significantly, we estimated the costs to run the 3.3B-parameter NLLB model using a single GPU on Google Colab. 
Details of our estimation method are in Appendix~\ref{sxn:det_est_nllb}. Table~\ref{tab:avg_cost} contains the average cost for each system across the languages we evaluated with it. 

\begin{table}[ht]
    \centering
    \small
    \begin{tabular}{lr}
        \textbf{model} &  \textbf{cost} \\
        \hline
        NLLB & \$0.09 \\
        ChatGPT (0-shot) & \$0.35 \\
        ChatGPT (5-shot) & \$1.32 \\
        Google & \$2.66 \\
        GPT-4 (5-shot) & \$25.93 \\
    \end{tabular}
    \caption{%\centering
    Estimated cost in USD to translate FLORES-200 \textit{devtest} ENG$\rightarrow$X with each system, averaged across all languages we evaluated with each
    }
    \label{tab:avg_cost}
\end{table}

\begin{figure}
    \centering
    \includegraphics[width=1\linewidth]{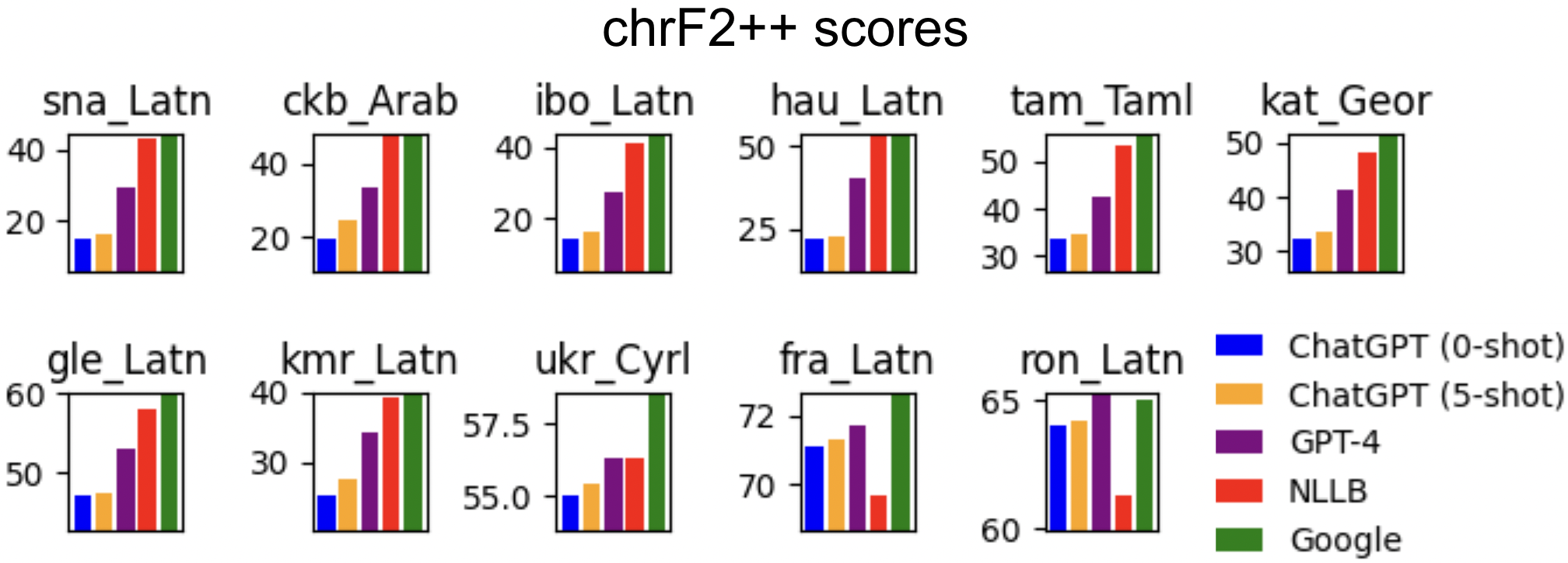}
    \includegraphics[width=1\linewidth]{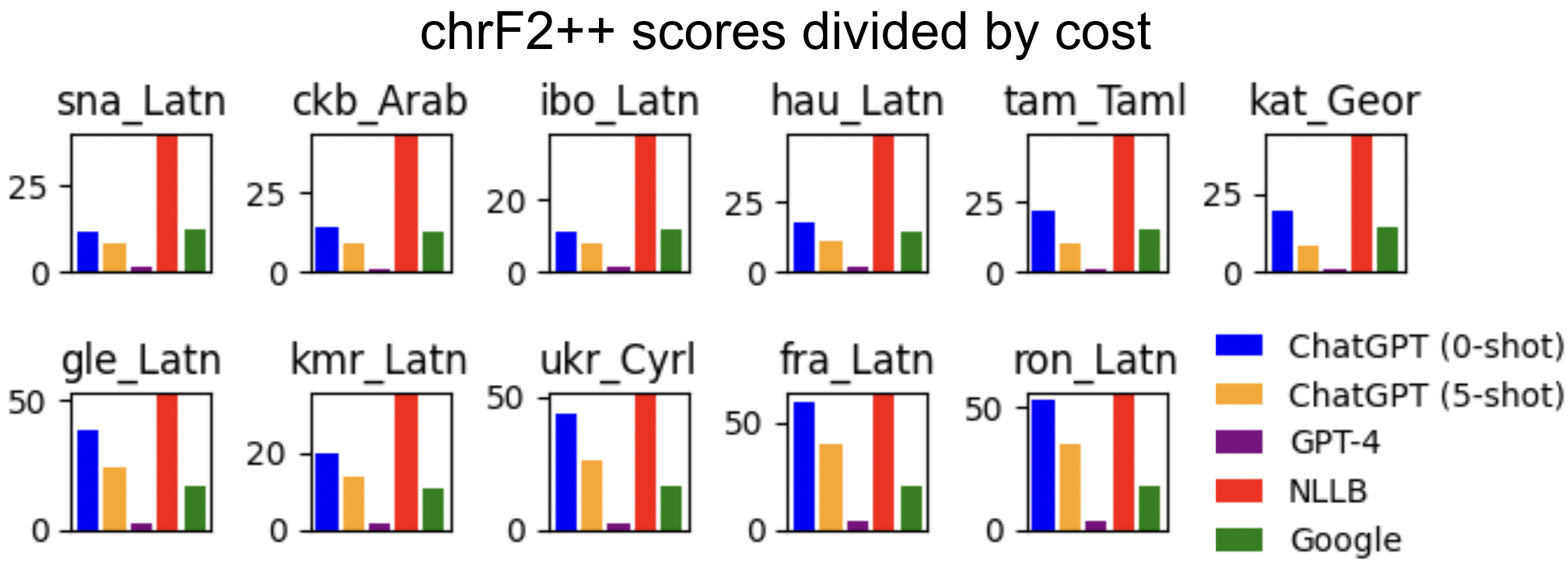}
    \caption{%\centering
    chrF scores for the 11 languages on which we evaluted all MT systems (top), followed by the same scores divided by the estimated cost of each system for each language (bottom)}
    \label{fig:11bars}
\end{figure}

Figure~\ref{fig:11bars} displays chrF scores for the 11 languages on which we evaluated all four MT systems (top), and the same scores divided by the approximate cost of each model (bottom). 
Bars for GPT-4 drop significantly in the bottom chart because of its high cost. 
Note from the top chart that Google Translate scores the best, but the bottom chart shows that NLLB has the best scores for its price. 
Zero-shot ChatGPT also tops five-shot in the bottom chart, suggesting that while few-shot prompts provide modest score improvements, they may not be worth the extra cost. 
See Appendix~\ref{sxn:app_bars} for fuller visualizations with all 203 languages. 

\section{Related Work}
\label{sxn:rw}

We are not the first researchers to explore LLM MT. 
However, most existing studies do not provide benchmarks for a large number or languages. 
\citet{wang2023document} studied GPT model discourse MT, but only for four languages. 
\citet{gao2023design} studied prompt engineering for GPT model MT, a helpful precursor to our work, but only for three languages.
\citet{moslem2023adaptive} probed the abilities of GPT models for adaptive and domain-appropriate MT  and term extraction, only including six languages in five directions. 
\citet{jiao2023chatgpt} produced MT benchmarks for ChatGPT and GPT-4, but only for five languages, none of them LRLs.\footnote{In this section, we define LRLs as languages having fewer than 1M Wikipedia pages.} 
They corroborated our findings that GPT models lag behind traditional MT models, but that GPT-4 outperforms ChatGPT. 
\citet{hendy2023good} explored 18 language pairs in a similar study, including four LRLs, but they focused more on MT performance across text domains, in-context learning, and reasoning than on multilingual benchmarks. 

In all the heretofore mentioned works combined, researchers explored only 18 languages, including five LRLs. 
This few-language approach does not address the needs of LLM users seeking to translate any languages other than the small few represented. 
In a work most comparable to our own, \citet{zhu2023multilingual} attempted to address this issue. 
They provided benchmarks comparing LLMs and traditional MT models across 102 languages, including 68 LRLs. 
Their results corroborate our own conclusions that LLMs lag behind traditional MT models, especially for LRLs. 
However, their analysis focuses primarily on few-shot learning and prompt engineering, including some topics somewhat removed from end user needs (such as the viability of nonsensical prompts in few-shot settings). 
Our work differs from existing studies in our focus on end users. 
We include more languages than any existing work (\textbf{204} languages, including \textbf{168} LRLs), to address the needs of various LRL communities. 
Our analysis suggests which language features predict LLM effectiveness, to help end users make hypotheses even about languages not represented in our study. 
We evaluate monetary costs, since they are a concern for LLM users. 

\section{Conclusion}

We provide benchmarks for LLM ENG$\rightarrow$X MT performance across 203 languages, with comparisons to state-of-the-art commercial and open-source MT models. 
For many HRLs, LLMs like ChatGPT perform competitively with these traditional models.  
But for LRLs, traditional MT remains dominant, despite LLMs' increased parameter size. 
Our decision-tree analysis reveals language features that predict ChatGPT's translation effectiveness relative to NLLB, finding that ChatGPT is especially disadvantaged for LRLs and African languages, and that the number of Wikipedia pages a language has is a strong predictor of ChatGPT's effectiveness in it. 
We present evidence that few-shot learning offers generally marginal improvements for ENG$\rightarrow$X MT, which may not justify its additional cost. 
We provide MT users with scores and cost estimates for four LLM and traditional MT systems, to help them determine which to use for their languages. 

Future work in this vein may include more translation directions (e.g. X$\rightarrow$ENG and non-English-centric), and human evaluation of LLM MT outputs to reveal trends along dimensions like fluency and accuracy. 
We open-source software and outputs of the models we evaluated on our \href{\github}{repository}.

\clearpage

\section*{Limitations}

We acknowledge limitations of using ChatGPT models for research. 
Since they are closed-source models, there is much we do not know about their architectural and training details, which can impact our understanding of their capabilities and biases.
For instance, OpenAI's implementation of mechanisms to prevent the generation of harmful or toxic content may inadvertently impact the quality of the model's output. 
This can be a concern when evaluating the reliability and accuracy of the results. 
OpenAI continuously updates and deprecates models behind the ChatGPT API, so our assessment may not be immaculate for future versions. 

While FLORES-200 is large and diverse, it is likely not representative of the vast array of languages worldwide. 
Some low-resource sets within FLORES-200 may contain noisy or corrupted data, potentially affecting the validity of the automatic metrics we employ in our reporting of scores. 
Additionally, FLORES-200 sets were translated from English Wikipedia. 
We avoided any X$\rightarrow$ENG translation directions, since it is likely that GPT models were trained on English Wikipedia. 
However, the semantic proximity of the other language sets to the original English source could potentially provide an advantage to these models in generating them. 
We also acknowledge the absence of non-English-centric translation directions from this study; we leave this for future work.

Lastly, the unavailability of semantic MT evaluation techniques like COMET \cite{rei2020comet} or BLEURT \cite{sellam2020bleurt} for LRLs hinders our ability to conduct comprehensive semantic evaluations and may leave some aspects of the translation quality unexplored. 
Human evaluation (which we leave for future work) may also reveal much in this area. 
These limitations surrounding model transparency, representative data, and evaluation should be taken into account when interpreting the findings of this work. 
Future studies may benefit from addressing these challenges to enhance the robustness and reliability of MT conclusions.

\section*{Ethics Statement}

The new prominance of LLMs in language technologies has numerous ethical implications. 
This study makes it apparent that even powerful LLMs like ChatGPT have significant limitations, such as an inability to translate a large number of low-resource languages. 
It also suggests that although these LLMs are trained on large and diverse data sets, they still have implicit biases, such as a clear disadvantage in MT for African languages. 
We hope to stress the importance of acknowledging and publicizing the limits and biases of these LLMs. 
This is especially relevant because a majority of LLM users may not be familiar or experienced with artificial intelligence (AI) engineering practices, and the commercial entities providing LLMs often have a monetary incentive to deliberately downplay the models' limitations. 
This can lead to unethical exploitation of users, who may attempt to use LLMs in applications where their limitations and biases can cause harm. 
Part of our goal in this work is to bring these discussions to the forefront of AI research. 
Ethical considerations like these should be a top concern for AI researchers, especially when many recent AI advancements are piloted by powerful commercial corporations. 

We hope also to acknowledge some of the ethical considerations involved in our own research. 
As we strive to develop improved open-source and accessible translation systems, it is essential to acknowledge that some language communities may have reservations about having their languages translated. 
Another crucial point is that utilizing the FLORES-200 test set in this research may inadvertently contribute to its incorporation into OpenAI's training data. 
OpenAI's current position is that API requests are not used for training \cite{openaipol}, but if this position were altered or disregarded, it could compromise the reliability of this test set for future GPT iterations. 
(This is a consideration for many commercial LLMs, though we only used OpenAI's in the current work.) 
This scenario has a  potential negative impact on the MT community, since many researchers depend on FLORES-200 and other MT benchmarks for large, diverse, high-quality data to conduct system comparisons.

\section*{Acknowledgements}

% comment out for anonymous submission
We thank Simran Khanuja for her help in running our Google Translate baseline and her general support. 
We also thank Alex Cabrera for his help developing our Zeno browser. 
This material is based on research sponsored in part by the Air Force Research Laboratory under agreement number FA8750-19-2-0200. The U.S. Government is authorized to reproduce and distribute reprints for Governmental purposes notwithstanding any copyright notation thereon. The views and conclusions contained herein are those of the authors and should not be interpreted as necessarily representing the official policies or endorsements, either expressed or implied, of the Air Force Research Laboratory or the U.S. Government. 
This work was also supported in part by the National Science Foundation under grant \#2040926, a grant from the Singapore Defence Science and Technology Agency.

\bibliography{anthology,custom}
\bibliographystyle{acl_natbib}

\appendix
\clearpage 

\section{Unabridged Result Table}
\label{sec:appendix}

In Table~\ref{tab:main_results} we report full results for 203 target languages in ENG$\rightarrow$X translation directions, across four MT systems: two LLMs (ChatGPT and GPT-4, with two $n$-shot settings for ChatGPT), one open-source encoder-decoder MT model (NLLB), and one commercial system (Google). 
We order in them in increasing order of performance, with zero-shot ChatGPT performing the worst and Google performing the best overall. 
We obtained scores for 203 target languages with ChatGPT, 201 with NLLB, 115 with Google Translate, and 20 with GPT-4. 
Our scores are spBLEU \cite{goyal-etal-2022-flores} using the SPM-200 tokenizer \cite{nllbteam2022language} and chrF2++ \cite{popovic-2017-chrf}. 
All results are also available on our \href{\github}{repository}, and interactive visualizations and histograms can be browsed on our Zeno \href{\zeno}{browser}.  

\section{Unabridged Bar Charts and Cost Estimation}
\label{sxn:app_bars}

\begin{figure*}[ht]
    \centering
    \includegraphics[width=1\linewidth]{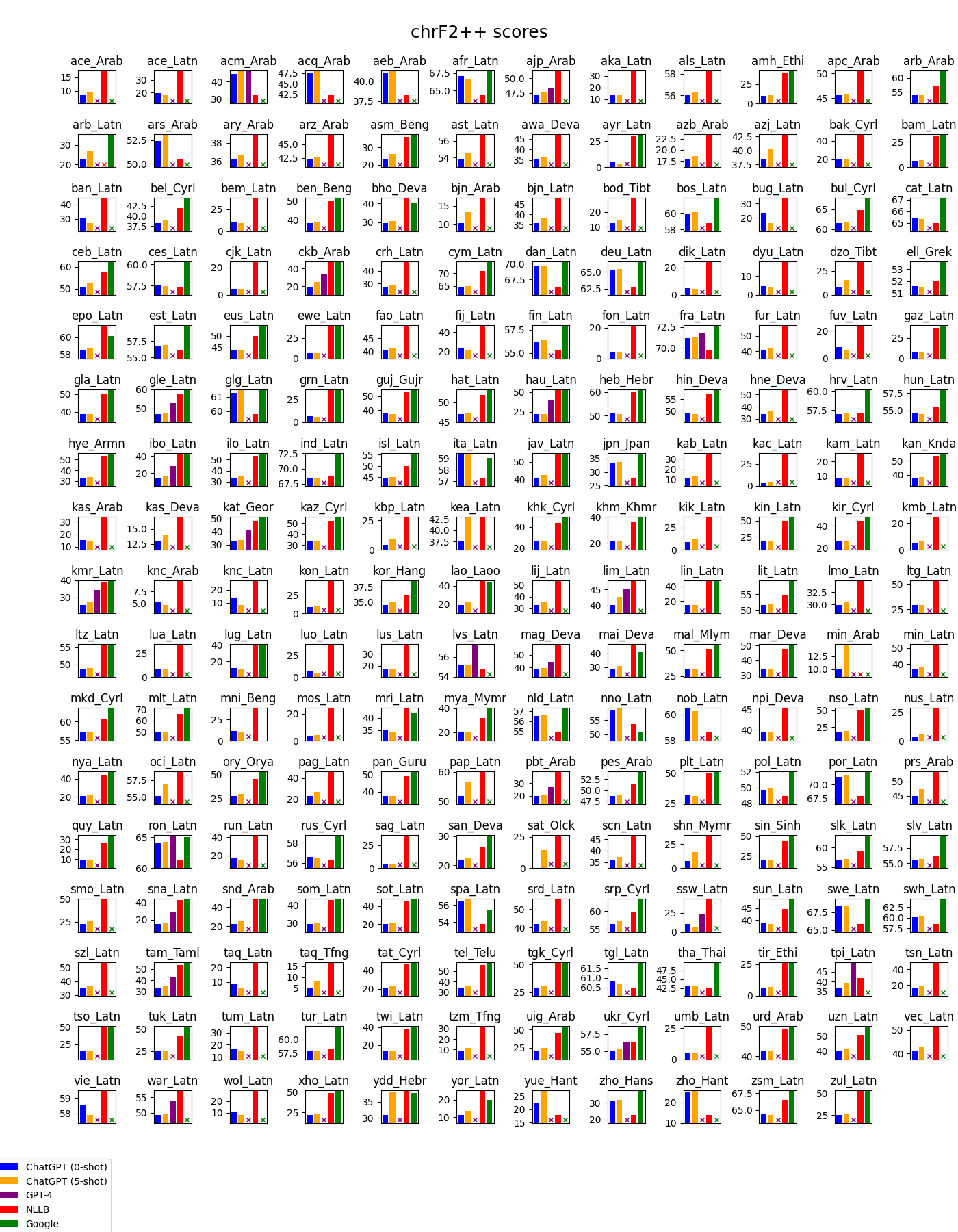}
    \caption{\centering
    chrF scores across all MT systems and languages 
    }
    \label{fig:big_chrF_bars}
\end{figure*}

\begin{figure*}
    \centering
    \includegraphics[width=1\linewidth]{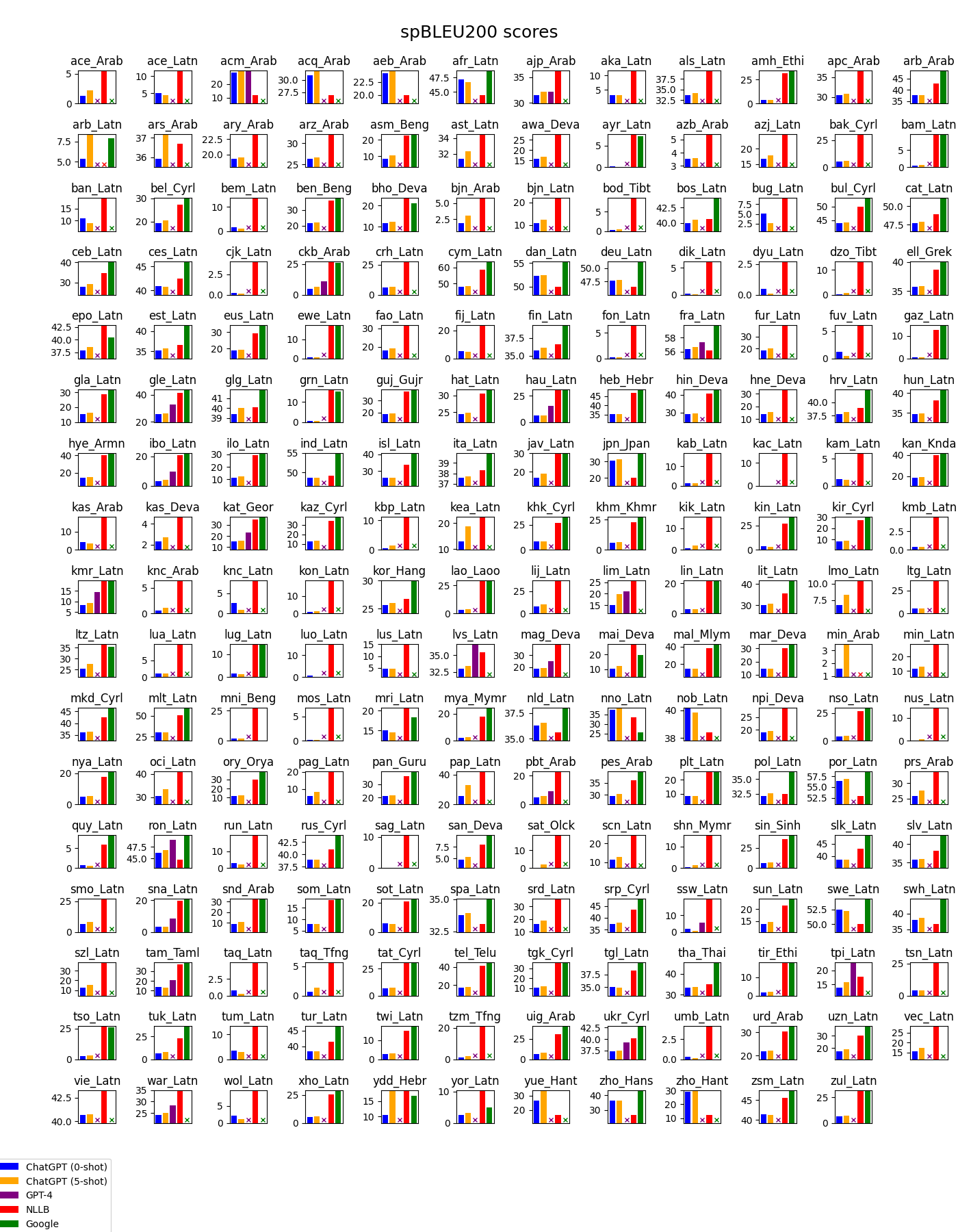}
    \caption{\centering
    BLEU scores across all MT systems and languages
    }
    \label{fig:big_bleu_bars}
\end{figure*}

\begin{figure*}
    \centering
    \includegraphics[width=1\linewidth]{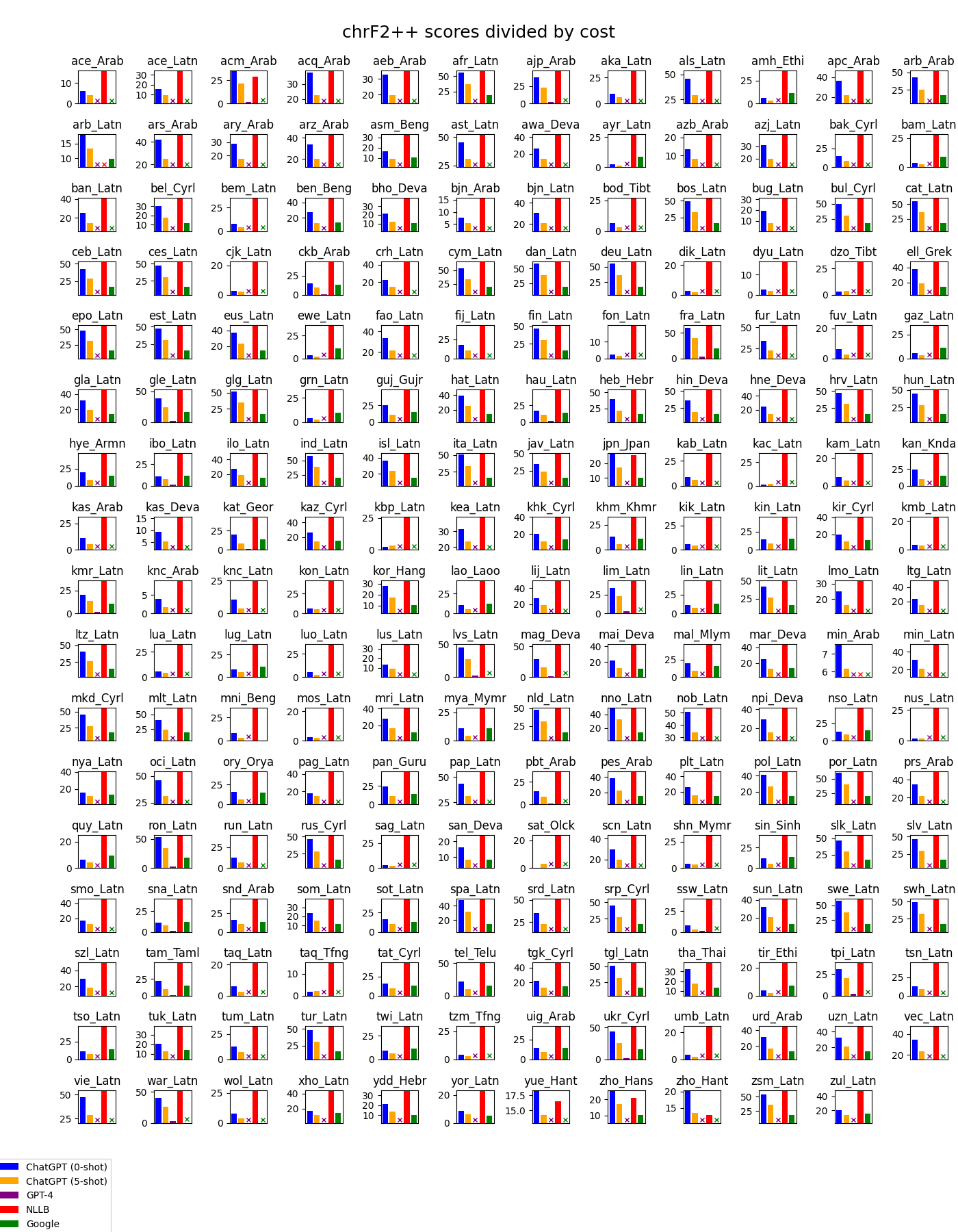}
    \caption{\centering
    chrF scores divided by the estimated cost of each MT system, across all MT systems and languages 
    }
    \label{fig:big_chrF_cost_bars}
\end{figure*}

\begin{figure*}
    \centering
    \includegraphics[width=1\linewidth]{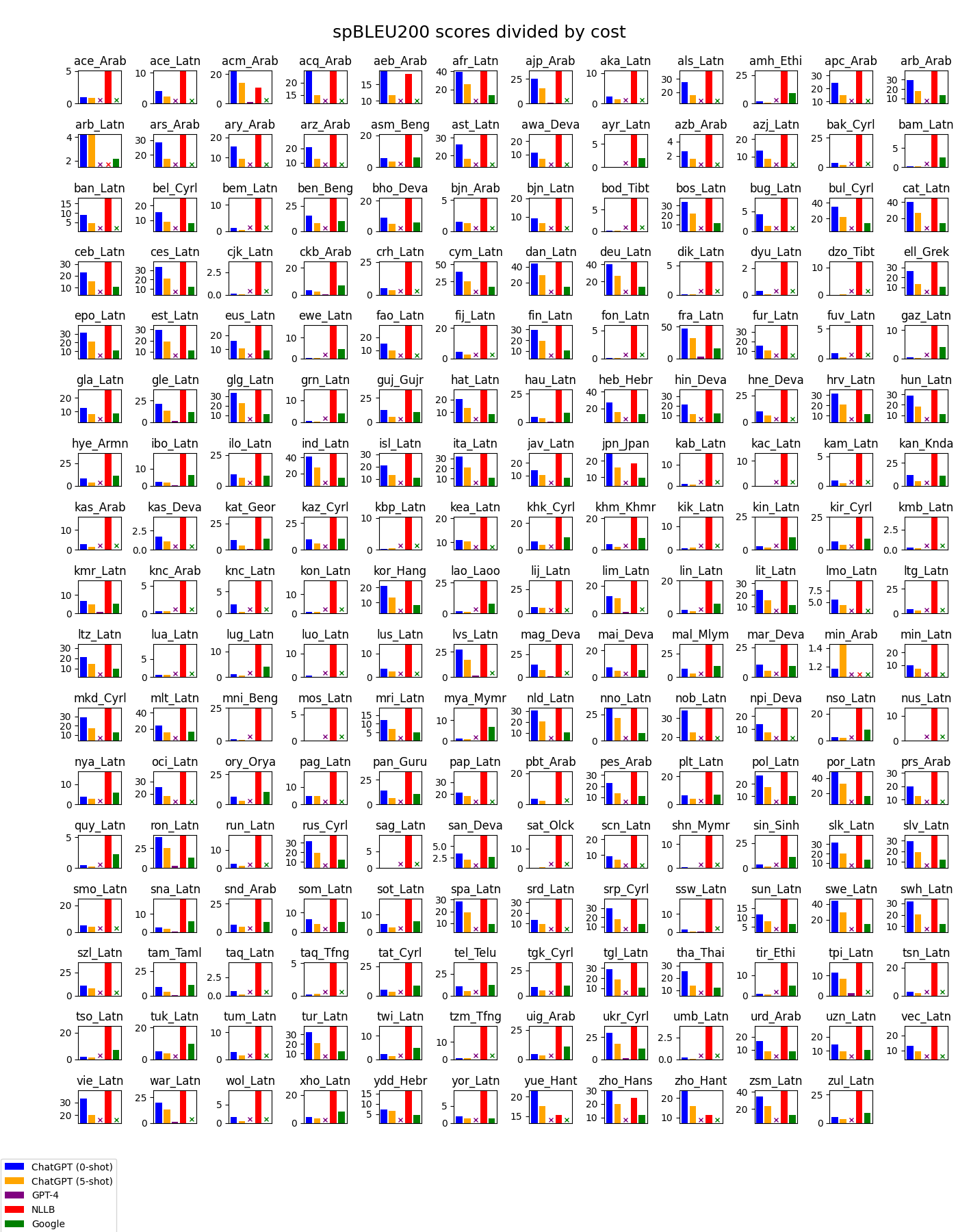}
    \caption{\centering
    BLEU scores divided by the estimated cost of each MT system, across all MT systems and languages
    }
    \label{fig:big_bleu_cost_bars}
\end{figure*}

See Figures~\ref{fig:big_chrF_bars} and \ref{fig:big_bleu_bars} for chrF and BLEU scores across all MT systems and languages. 
Google Translate and NLLB are generally the best performers in both metrics, though GPT-4 and ChatGPT are occasionally best. 
An ``x'' indicates where we did not evaluate one of the systems for a language. 
Figures~\ref{fig:big_chrF_cost_bars} and \ref{fig:big_bleu_cost_bars} display chrF and BLEU scores divided by the estimated cost of each MT system. The cost value is measured as the amount in USD that it would cost to translate the entire FLORES-200 \textit{devtest} set for each language. 

These visualizations are also available on our \href{\github}{repository}. 
(Also see our Zeno \href{\zeno}{browser} for interactive visualizations of our results.) 
We also include cost estimates and scores divided thereby for all languages and MT systems in Table~\ref{tab:full_costs}. 
We exclude cost estimates by language for NLLB and Google because there is very little variation between languages. 
Our estimated cost of translating FLORES-200 \textit{devtest} ENG$\rightarrow$ is approximately \$0.09 for every target language. 
And the respective estimate for Google Translate is roughly \$2.66 regardless of the target language, since Google's API only charges for input characters. 

\subsection{Details about estimating NLLB cost}
\label{sxn:det_est_nllb}

To estimate the cost of running NLLB's 3.3B-parameter model for translation, we used one GPU from Google Colab to translate the full FLORES-200 \textit{devtest} set from English into six languages representing six high- and low-resource scripts--Burmese (\texttt{mya\_Mymr}), Simplified Chinese (\texttt{zho\_Hans}), Standard Arabic (\texttt{arb\_Arab}), Hindi (\texttt{hin\_Deva}), Armenian (\texttt{hye\_Armn}), and French (\texttt{fra\_Latn})--and measured the time for each. 
We assumed that runtime $t$ is determined by an equation with unknown coefficients $x_1$, $x_2$, and $x_3$:

\begin{equation}
\label{eqn:lin}
    t = x_1n_{input} + x_2n_{output} + x_3
\end{equation}

where $n_{input}$ represents the number of input tokens and $n_{output}$ is the number of output tokens. 
In this case, $x_1$ represents the rate at which the encoder processes input tokens, $x_2$ represents the rate at which the decoder undergoes inference, and $x_3$ is the amount of time to perform all other computations, independent of the number of tokens. We estimated $x_1$, $x_2$, and $x_3$ via a least-squares solution to the linear system defined by the six languages for which we obtained runtime $t$:

$$
\begin{bmatrix}
n_{input} & n_{output}(\texttt{mya}) & 1\\
n_{input} & n_{output}(\texttt{zho}) & 1\\
n_{input} & n_{output}(\texttt{arb}) & 1\\
n_{input} & n_{output}(\texttt{hin}) & 1\\
n_{input} & n_{output}(\texttt{hye}) & 1\\
n_{input} & n_{output}(\texttt{fra}) & 1\\
\end{bmatrix}
\begin{bmatrix}
x_1\\
x_2 \\
x_3 \\
\end{bmatrix} = 
\begin{bmatrix}
t_{\texttt{mya}} \\
t_{\texttt{zho}} \\
t_{\texttt{arb}} \\
t_{\texttt{hin}} \\
t_{\texttt{hye}} \\
t_{\texttt{fra}} \\
\end{bmatrix}
$$

where $n_{input}$ is the number of tokens in the English \textit{devtest} set, and $n_{output}$ for each language is the number of tokens in the NLLB-MOE model output provided by \citet{nllbteam2022language}. (We used the same tokenizer that we had used for GPT model cost estimation, for simplicity.) After estimating $x_1$, $x_2$, and $x_3$, we used them in Equation~\ref{eqn:lin} to estimate $t$ values for all 201 languages for which we obtained NLLB MT scores. We then used Google Colab's estimated rate of \$0.35/hour for use of one GPU to estimate costs for each language.

\section{Visualizations Comparing ChatGPT and NLLB}
\label{sxn:app_hexbins}

\begin{figure*}
    \centering
    % \top
    \includegraphics[width=0.45\linewidth]{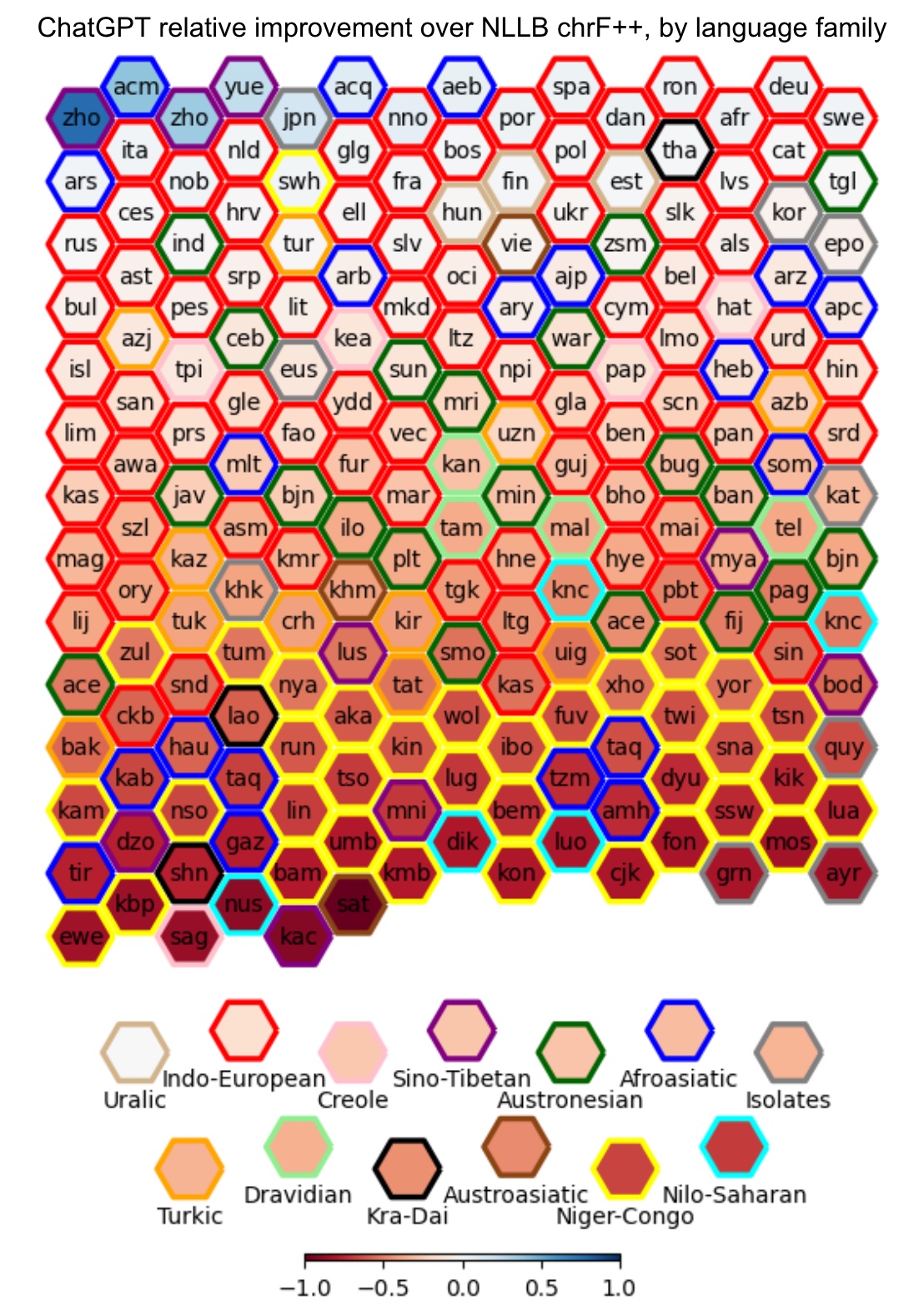}
    \includegraphics[width=0.45\linewidth]{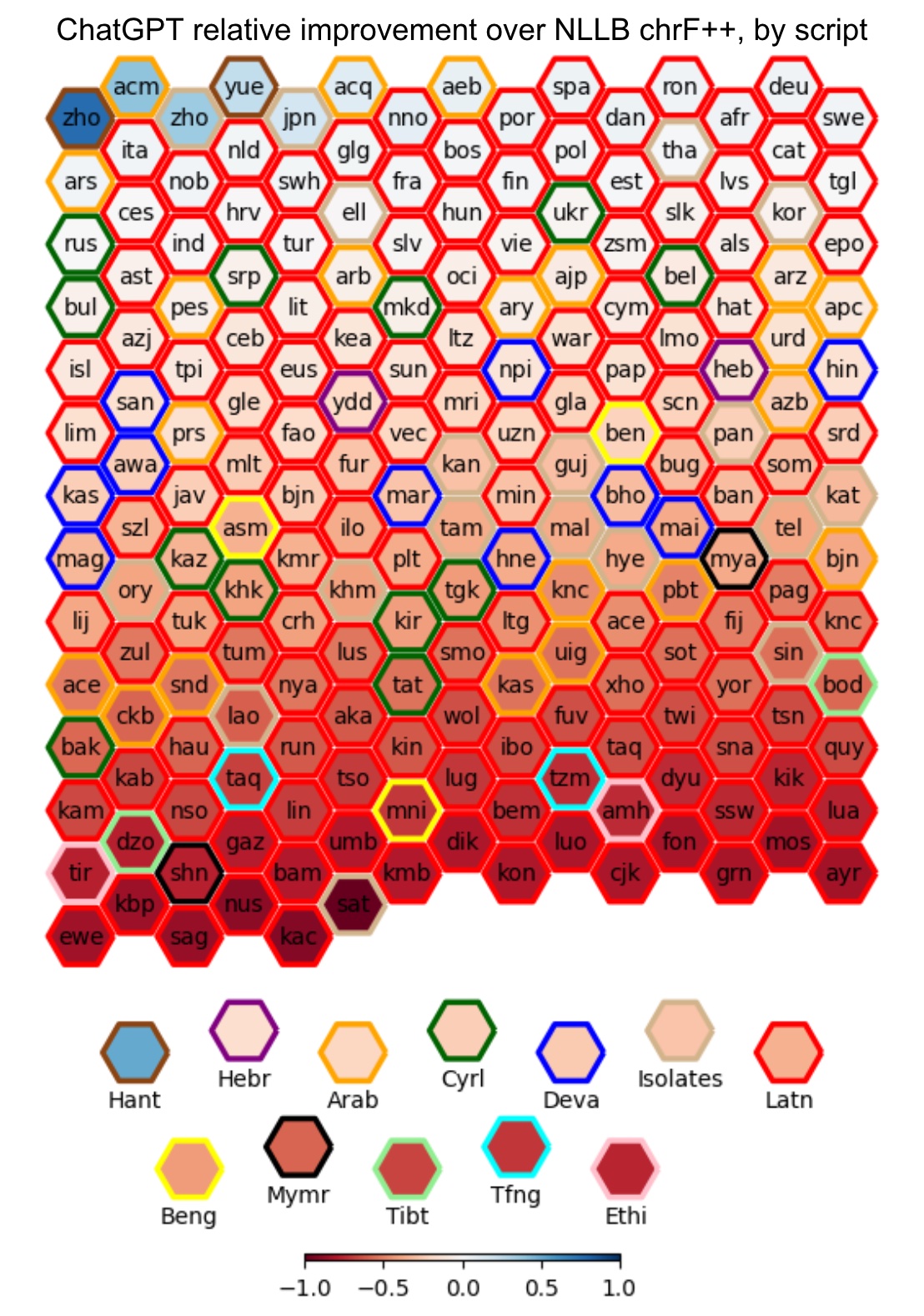}
    \includegraphics[width=0.45\linewidth]{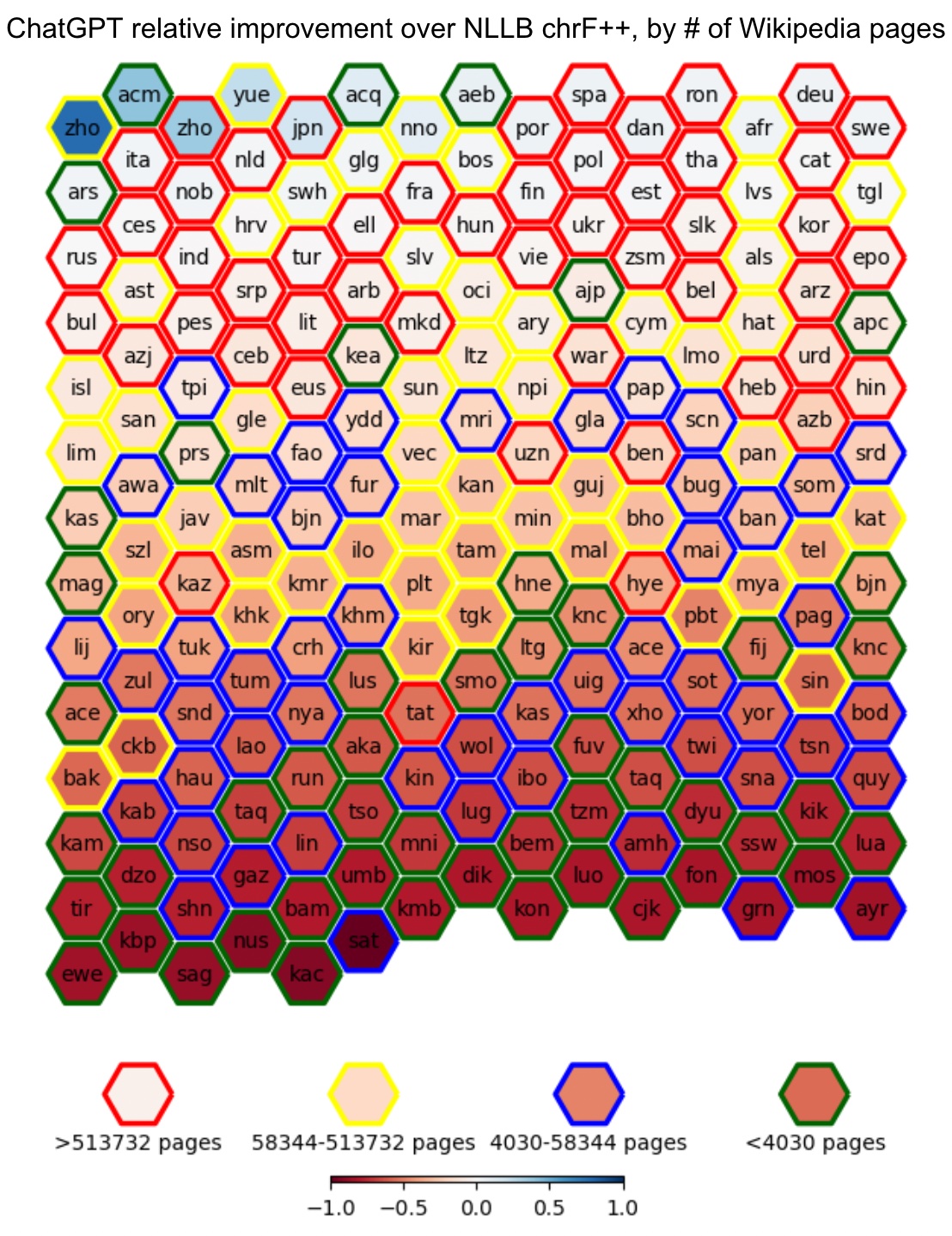}
    \caption{%\centering
    ChatGPT \textit{relative improvement} over NLLB chrF (color scale), with languages organized by family, script, and number of Wikipedia pages (divided in quartiles). Hexagons (one per language) are displayed in descending order across rows, with the highest ChatGPT relative improvement over NLLB chrF2++ at the top left, and the lowest at the bottom right. Group hexagons at the bottom of each plot display the average color for each group and are organized in like manner.}
    \label{fig:bighexplots}
\end{figure*}

\begin{figure*}
    \centering
    % \top
    \includegraphics[width=1\linewidth]{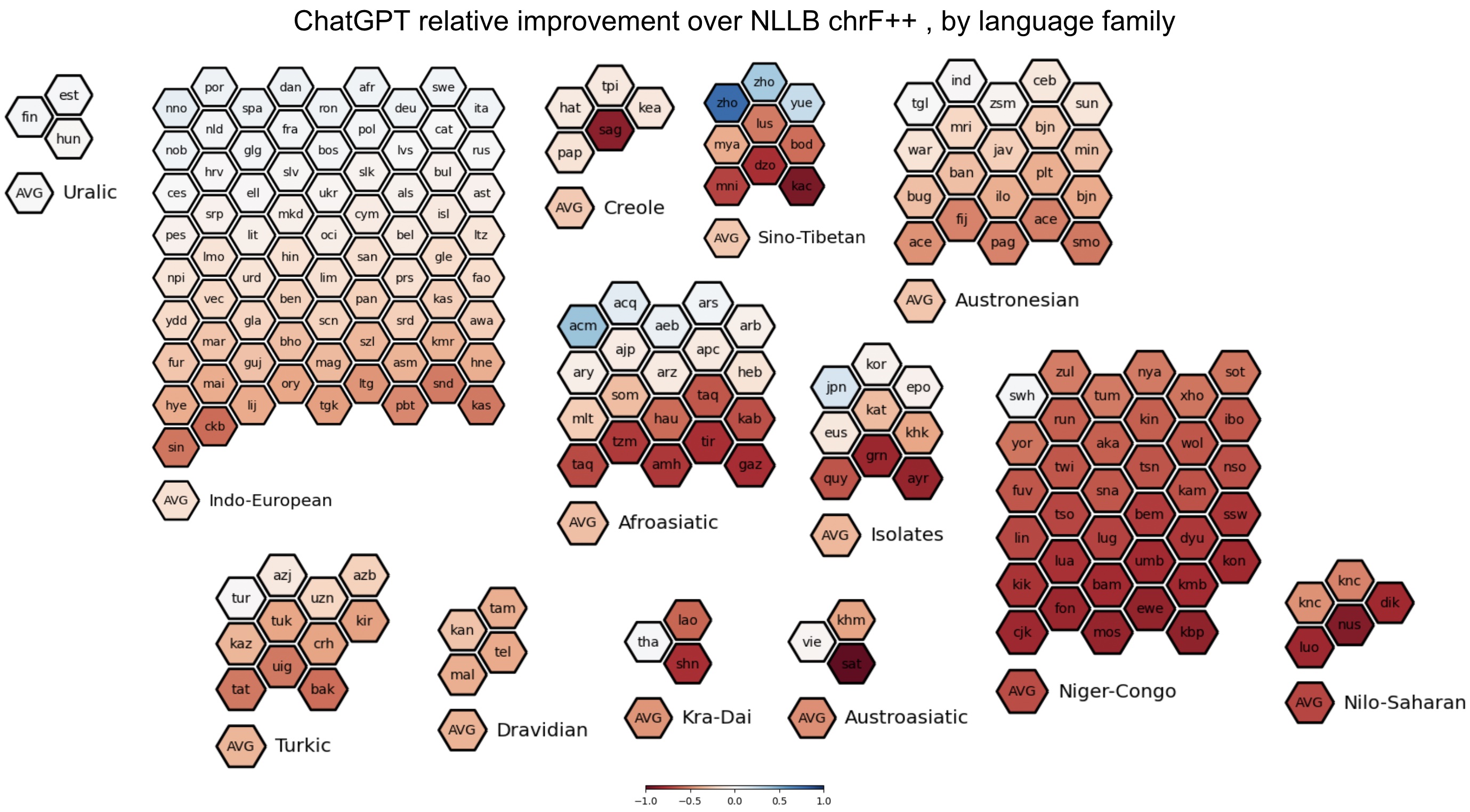}\\
    % \rule{\linewidth}{0.5pt}
    \includegraphics[width=1\linewidth]{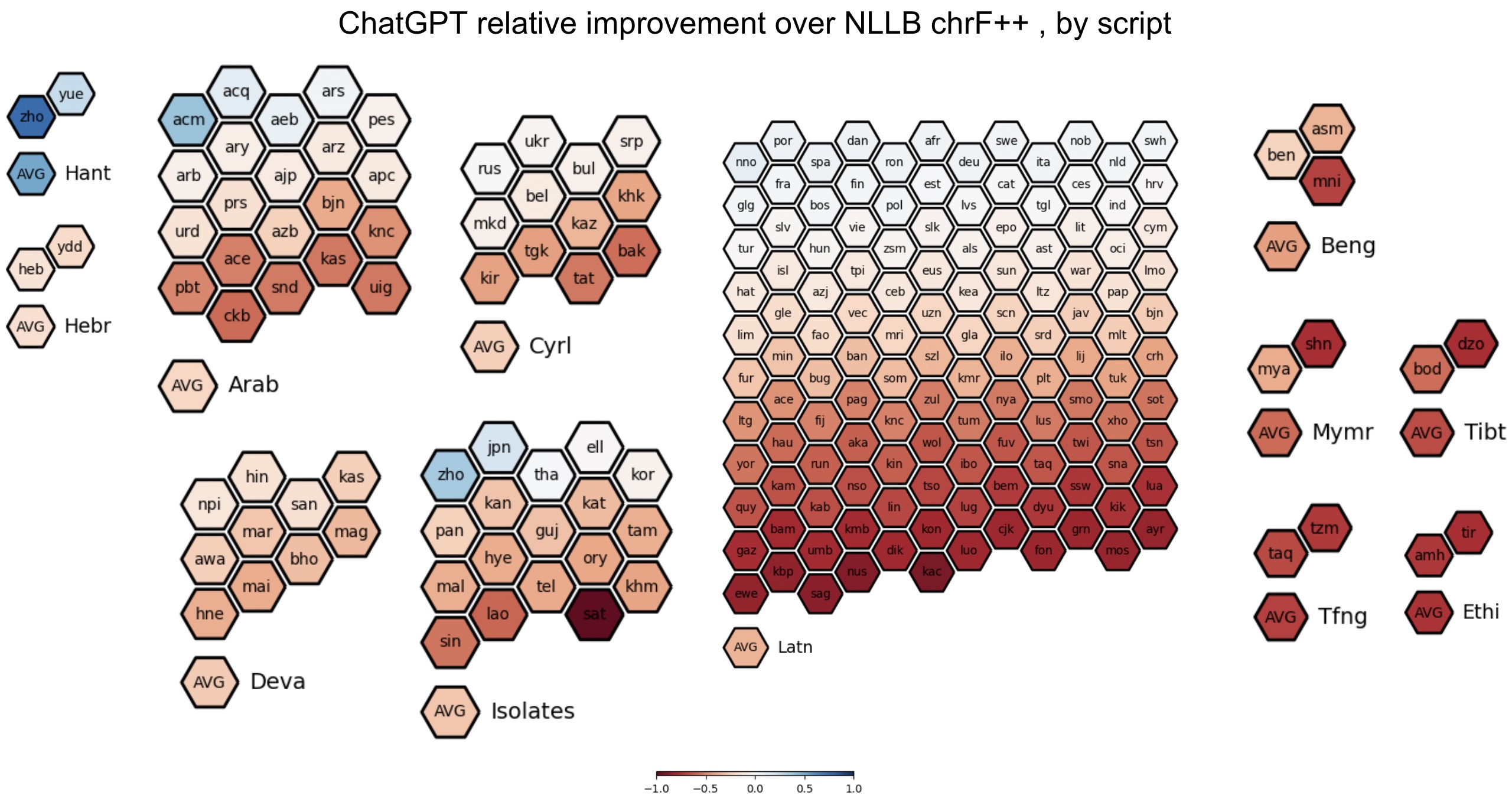}\\
    % \rule{\linewidth}{0.5pt}
    \includegraphics[width=1\linewidth]{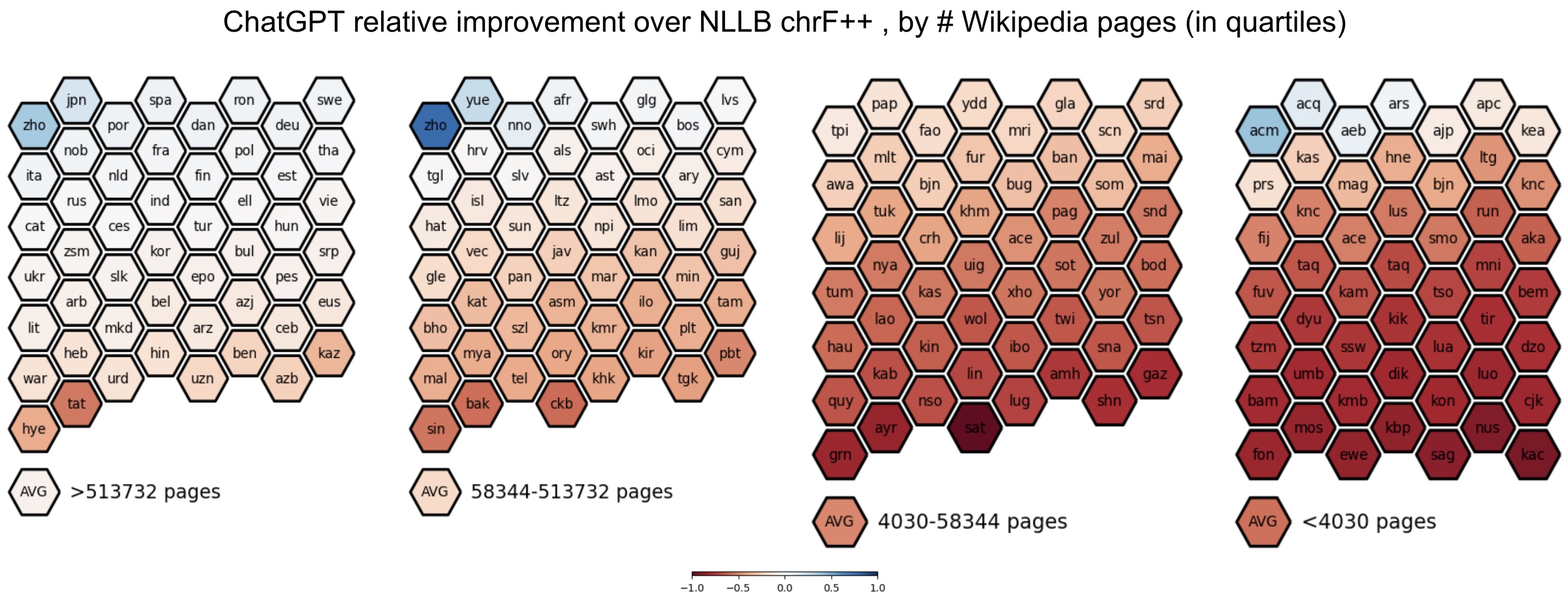}
    \caption{%\centering
    Alternative visualizations to those in Figure~\ref{fig:bighexplots}. Groups and languages are organized the same here: from top left to bottom right in descending order of the ChatGPT \textit{relative improvement} over NLLB (using averages for the groups).
    }
    \label{fig:composed_hexplots}
\end{figure*}

See Figures \ref{fig:bighexplots} and \ref{fig:composed_hexplots}. 
They are also posted on our \href{\github}{repository}. 
(Also see our Zeno \href{\zeno}{browser} for interactive visualizations of our results.)

\section{Estimating Wikipedia Page Counts}

As mentioned in \S\ref{sxn:fewshot}, we used the "Total pages" count from \url{https://en.wikipedia.org/wiki/List_of_Wikipedias}, accessed 7 August 2023, as a proxy for the resource level of a language (refered to as \texttt{wiki\_ct} in \S\ref{sxn:lang_feats}). We had to make some decisions regarding macrolanguage and microlanguage matches when making these estimates. Many of the languages in FLORES-200 \citep{nllbteam2022language} are in fact microlanguages of a macrolangauge not included in the dataset. In some cases this microlanguage was did not have a listed Wikipedia page count, so we used the macrolanguage page count instead. Table~\ref{tab:macros1} lists all the languages for which we used the Wikipedia page count of a macrolanguage (with a different ISO 639-3 code), based on our best judgment. In every case this was because the FLORES-200 microlanguage was not listed.

There were also cases where we decided to list zero for a microlanguage's \texttt{wiki\_ct}, even if its macrolanguage was listed with a nonzero number of pages. This was in cases where we could reasonably assume that the macrolanguage's Wikipedia pages were likely (either all or predominantly) in another microlanguage or dialect. We list the languages that we considered in this manner in Table~\ref{tab:macros2}. 

We also made some decisions regarding \texttt{wiki\_ct} assignment based on the script of a language. We recorded zero Wikipedia pages for \texttt{kas\_Deva} and 13,210 for \texttt{kas\_Arab} (all of the Kashmiri pages) because a majority of Kashmiri pages seem to be in Perso-Arabic script. (There may be a few in Devanagari, but we simplify by assuming none are.) We also recorded zero pages for \texttt{mni\_Beng} because, although Wikipedia has pages in Meitei, they appear to be in the Meitei \texttt{Mtei} script, not Bengali \texttt{Beng}. Lastly, we assigned Wikipedia's count for `Classical Chinese' (\texttt{zh-classical}) to \texttt{zho\_Hant} and its count for `Chinese' to \texttt{zho\_Hans} (though it is possible that some of the `Chinese' pages may be in the Traditional Chinese (\texttt{Hant}) script).

In all other cases, if a language did not have a listed number of Wikipedia pages, we took this to mean it had zero.

\onecolumn

\begin{center}  
\small
\begin{longtable}{l|rrrrr|rrrrr}
\caption{%\centering 
BLEU and chrF results on ENG$\rightarrow$X directions. ``0-shot" and ``5-shot" are ChatGPT with zero- and five-shot settings, respectively. ``NLLB" is the NLLB-MOE model, and ``Google" is Google Translate. We used five-shot settings only for GPT-4. Models are listed in order of their effectiveness in MT (with zero-shot ChatGPT performing the worst and Google Translate performing the best).} \\
% \hline
\textbf{Language  }    & \multicolumn{5}{c|}{\textbf{spBLEU200}}                             & \multicolumn{5}{c}{\textbf{chrF2++}}                                                                                               \\
          & \multicolumn{1}{l}{0-shot} & \multicolumn{1}{l}{5-shot} & \multicolumn{1}{l}{GPT-4} & \multicolumn{1}{l}{NLLB}     & \multicolumn{1}{l|}{Google} & \multicolumn{1}{l}{0-shot} & \multicolumn{1}{l}{5-shot} & \multicolumn{1}{l}{GPT-4} & \multicolumn{1}{l}{NLLB}     & \multicolumn{1}{l}{Google} \\
          
\hline
\endfirsthead

\multicolumn{11}{c}%
{ \tablename\ \thetable{} -- continued from previous page}\\
\textbf{Language  }    & \multicolumn{5}{c|}{\textbf{spBLEU200}}                             & \multicolumn{5}{c}{\textbf{chrF2++}}                                                                                               \\
          & \multicolumn{1}{l}{0-shot} & \multicolumn{1}{l}{5-shot} & \multicolumn{1}{l}{GPT-4} & \multicolumn{1}{l}{NLLB}     & \multicolumn{1}{l|}{Google} & \multicolumn{1}{l}{0-shot} & \multicolumn{1}{l}{5-shot} & \multicolumn{1}{l}{GPT-4} & \multicolumn{1}{l}{NLLB}     & \multicolumn{1}{l}{Google} \\
\hline 
\endhead

\hline \multicolumn{11}{|c|}{{Continued on next page}} \\ \hline
\endfoot
\hline
\endlastfoot

ace\_Arab & 1.3                   & 2.3                   & --         & 5.5       & --          & 8.4                   & 9.8                   & --         & 17.4      & --          \\
ace\_Latn & 5.0                   & 4.3                   & --         & 11.6      & --          & 19.8                  & 18.0                  & --         & 37.1      & --          \\
acm\_Arab & 28.2                  & 29.6                  & 29.5       & 11.8      & --          & 44.7                  & 46.1                  & 46.5       & 31.9      & --          \\
acq\_Arab & 30.9                  & 31.9                  & --         & 26.9      & --          & 47.5                  & 48.1                  & --         & 42.2      & --          \\
aeb\_Arab & 24.2                  & 24.7                  & --         & 19.9      & --          & 41.0                  & 41.3                  & --         & 38.2      & --          \\
afr\_Latn & 47.2                  & 46.7                  & --         & 44.4      & 48.7        & 67.0                  & 66.7                  & --         & 64.3      & 67.8        \\
ajp\_Arab & 31.5                  & 32.2                  & 32.2       & 36.3      & --          & 47.1                  & 47.5                  & 48.4       & 51.3      & --          \\
aka\_Latn & 3.2                   & 3.1                   & --         & 11.7      & --          & 13.3                  & 13.8                  & --         & 34.5      & --          \\
\end{longtable} \end{center} 

\clearpage

\begin{table*}[]
    \centering
    \small
\begin{tabular}{ll}
        \textbf{FLORES lang.} &  \textbf{substitution for \texttt{wiki\_ct}} \\
        \hline
        \texttt{arb} & Used macrolanguage `Arabic' (\texttt{ara}) because `Standard Arabic' (\texttt{arb}) not present \\
        \texttt{bho} & Used macrolanguage `Bihari' (\texttt{bih}) because `Bhojpuri' (\texttt{bho}) not present \\
        \texttt{dik} & Used macrolanguage `Dinka' (\texttt{din}) because `Southwestern Dinka' (\texttt{dik}) not present \\
        \texttt{fuv} & Used macrolanguage `Fula' (\texttt{ful}) because "Nigerian Fulfulde" (\texttt{fuv}) not present \\
        \texttt{knc} & Used macrolanguage `Kanuri' (\texttt{kau}) because `Central Kanuri' (\texttt{knc}) not present \\
        \texttt{lvs} & Used macrolanguage `Latvian' (\texttt{lav}) because `Standard Latvian' (\texttt{lvs}) not present \\
        \texttt{plt} & Used macrolanguage `Malagasy' (\texttt{mlg}) because `Plateau Malagasy' (\texttt{plt}) not present \\
\end{tabular}
    \caption{FLORES-200 languages for which we used the Wikipedia page count associated with a macrolanguage of another ISO 639-3 code}
    \label{tab:macros1}
\end{table*}

\begin{table*}[]
    \centering
    \small
    \begin{tabular}{ll}
        \textbf{FLORES lang.} &  \textbf{reason for assigning \texttt{wiki\_ct} = 0} \\
        \hline
        \texttt{acm} & Macrolanguage `Arabic' (\texttt{ara}) appears to be in `Standard Arabic' (\texttt{arb}), not `Mesopotamian Arabic' (\texttt{acm}) \\
        \texttt{acq} & Macrolanguage `Arabic' (\texttt{ara}) appears to be in `Standard Arabic' (\texttt{arb}), not `Tai'izzi Arabic' (\texttt{acq}) \\
        \texttt{aeb} & Macrolanguage `Arabic' (\texttt{ara}) appears to be in `Standard Arabic' (\texttt{arb}), not `Tunisian Arabic' (\texttt{aeb}) \\
        \texttt{ajp} & Macrolanguage `Arabic' (\texttt{ara}) appears to be in `Standard Arabic' (\texttt{arb}), not `South Levantine Arabic' (\texttt{ajp}) \\
        \texttt{apc} & Macrolanguage `Arabic' (\texttt{ara}) appears to be in `Standard Arabic' (\texttt{arb}), not `North Levantine Arabic' (\texttt{apc}) \\
        \texttt{ars} & Macrolanguage `Arabic' (\texttt{ara}) appears to be in `Standard Arabic' (\texttt{arb}), not `Najdi Arabic' (\texttt{ars}) \\
        \texttt{mag} & Macrolanguage `Bihari' (\texttt{bih}) appears to be in `Bhojpuri' (\texttt{bho}), not `Magahi' (\texttt{mag}) \\
        \texttt{prs} & Macrolanguage `Persian' (\texttt{fas}) appears to be in `Western Persian' (\texttt{pes}), not `Dari' (\texttt{prs}) 
        
    \end{tabular}
    \caption{FLORES-200 languages for which we used assigned \texttt{wiki\_ct} to be zero, despite the existence of Wikipedia pages in a corresponding macrolanguage}
    \label{tab:macros2}
\end{table*}

\begin{center}  
\scriptsize
\begin{longtable}{l|rrrrr|rrrrr|rrrHH}
\caption{%\centering 
Estimated costs in USD to translate the FLORES-200 \textit{devtest} set ENG$\rightarrow$X for each targe language and MT system, along with BLEU and chrF scores divided by the cost estimates, where applicable. The cost is roughly \$0.09 for NLLB and \$2.66 for Google Translate for all target languages.
\label{tab:full_costs} 
} \\
% \hline
\textbf{Lang.}    & \multicolumn{5}{c|}{\textbf{spBLEU200/cost}}                             & \multicolumn{5}{c|}{\textbf{chrF2++/cost}}                             & \multicolumn{3}{c}{\textbf{cost estimate} (USD\$)}                                                                                               \\
          & \multicolumn{1}{l}{0-shot} & \multicolumn{1}{l}{5-shot} & \multicolumn{1}{l}{GPT-4} & \multicolumn{1}{l}{NLLB}     & \multicolumn{1}{l|}{Google} & \multicolumn{1}{l}{0-shot} & \multicolumn{1}{l}{5-shot} & \multicolumn{1}{l}{GPT-4} & \multicolumn{1}{l}{NLLB}     & \multicolumn{1}{l|}{Google}  & \multicolumn{1}{l}{0-shot} & \multicolumn{1}{l}{5-shot} & \multicolumn{1}{l}{GPT-4} \\ 
          
\hline
\endfirsthead

\multicolumn{16}{c}%
{ \tablename\ \thetable{} -- continued from previous page}\\
\textbf{Lang.}    & \multicolumn{5}{c|}{\textbf{spBLEU200/cost}}                             & \multicolumn{5}{c|}{\textbf{chrF2++/cost}}                             & \multicolumn{3}{c}{\textbf{cost estimate} (USD\$)}                                                                                               \\
          & \multicolumn{1}{l}{0-shot} & \multicolumn{1}{l}{5-shot} & \multicolumn{1}{l}{GPT-4} & \multicolumn{1}{l}{NLLB}     & \multicolumn{1}{l|}{Google} & \multicolumn{1}{l}{0-shot} & \multicolumn{1}{l}{5-shot} & \multicolumn{1}{l}{GPT-4} & \multicolumn{1}{l}{NLLB}     & \multicolumn{1}{l|}{Google}  & \multicolumn{1}{l}{0-shot} & \multicolumn{1}{l}{5-shot} & \multicolumn{1}{l}{GPT-4} \\ 
          
\hline
\endhead

\hline \multicolumn{16}{|c|}{{Continued on next page}} \\ \hline
\endfoot
\hline
\endlastfoot

ace\_Arab & 0.9                        & 0.9                        & --              & 5.1            & --               & 6.3                        & 4.0                        & --              & 16.0           & --               & 0.3                   & 1.5                   & 29.0       & 0.1       & 2.7         \\
ace\_Latn & 4.0                        & 2.2                        & --              & 10.7           & --               & 15.8                       & 9.1                        & --              & 34.2           & --               & 0.3                   & 1.0                   & 18.9       & 0.1       & 2.7         \\
acm\_Arab & 22.4                       & 13.9                       & 1.2             & 10.8           & --               & 35.6                       & 21.6                       & 1.8             & 29.3           & --               & 0.3                   & 1.1                   & 24.3       & 0.1       & 2.7         \\
acq\_Arab & 24.6                       & 14.9                       & --              & 24.8           & --               & 37.7                       & 22.4                       & --              & 38.9           & --               & 0.3                   & 1.1                   & 24.6       & 0.1       & 2.7         \\
aeb\_Arab & 19.3                       & 11.6                       & --              & 18.3           & --               & 32.7                       & 19.4                       & --              & 35.2           & --               & 0.3                   & 1.1                   & 24.1       & 0.1       & 2.7         \\
afr\_Latn & 39.5                       & 25.8                       & --              & 41.0           & 13.3             & 56.2                       & 36.8                       & --              & 59.4           & 18.5             & 0.2                   & 0.8                   & 17.1       & 0.1       & 2.7         \\
\end{longtable} 
\end{center} 

\clearpage 

\end{document}